\documentclass{article}

\usepackage{arxiv}

\usepackage[utf8]{inputenc} % allow utf-8 input
\usepackage[T1]{fontenc}    % use 8-bit T1 fonts
\usepackage{hyperref}       % hyperlinks
\usepackage{url}            % simple URL typesetting
\usepackage{booktabs}       % professional-quality tables
\usepackage{amsfonts}       % blackboard math symbols
\usepackage{nicefrac}       % compact symbols for 1/2, etc.
\usepackage{microtype}      % microtypography
\usepackage{lipsum}		% Can be removed after putting your text content
\usepackage{natbib}
\usepackage{doi}
\usepackage{graphicx}
\usepackage{subcaption}
\usepackage{multirow}
\usepackage{makecell}
\usepackage{xcolor}
\usepackage{adjustbox}

\usepackage{amsmath}
\usepackage{amssymb}
\usepackage{mathtools}
\usepackage{amsthm}
\usepackage[capitalize,noabbrev]{cleveref}

\title{i-PhysGaussian: Implicit Physical Simulation for 3D Gaussian Splatting}

\author{
  Yicheng Cao$^{1}$,
  Zhuo Huang$^{1}$,
  Yu Yao$^{1}$,
  Yiming Ying$^{2}$,
  Daoyi Dong$^{3}$,
  Tongliang Liu$^{1}$\\[1ex]
  \small{$^1$Sydney AI Centre, The University of Sydney;}
  \small{$^2$School of Mathematics and Statistics, University of Sydney;} \\
  \small{$^3$Australian Artificial Intelligence Institute, University of Technology Sydney}
}
\date{}

\renewcommand{\shorttitle}{i-PhysGaussian: Implicit Physical Simulation for 3D Gaussian Splatting}

\begin{document}
\maketitle

\begin{abstract}
    Physical simulation predicts future states of objects based on material properties and external loads, enabling blueprints for both Industry and Engineering to conduct risk management. Current 3D reconstruction-based simulators typically rely on explicit, step-wise updates, which are sensitive to step time and suffer from rapid accuracy degradation under complicated scenarios, such as high-stiffness materials or quasi-static movement. To address this, we introduce i-PhysGaussian, a framework that couples 3D Gaussian Splatting (3DGS) with an implicit Material Point Method (MPM) integrator. Unlike explicit methods, our solution obtains an end-of-step state by minimizing a momentum-balance residual through implicit Newton-type optimization with a GMRES solver. This formulation significantly reduces time-step sensitivity and ensures physical consistency. Our results demonstrate that i-PhysGaussian maintains stability at up to 20$\times$ larger time steps than explicit baselines, preserving structural coherence and smooth motion even in complex dynamic transitions.
\end{abstract}

\section{Introduction}
\label{sec:intro}

Recent advances in dynamic scene generation have enabled realistic spatiotemporal content for applications such as robotic manipulation~\cite{Lee_2025_CVPR, Wang_2024_ICML_RoboGen, Katara_2023_Gen2Sim, Zhou_2024_RoboDreamer} and video synthesis~\cite{Ren_2025_CVPR_GEN3C, Yu_2024_4Real, Chu_2024_NeurIPS_DreamScene4D, Huang_2025_Voyager}. Within this broader landscape, physics-based dynamic simulation seeks to generate temporally coherent motion that strictly respects physical laws by coupling neural scene representations with classical mechanics solvers. This capability is critical for forecasting object behavior under unseen loading conditions~\cite{jayaraman2018learning,huang2023harnessing,huang2024winning,li2024dynamic}, enabling long-horizon analysis and design with high fidelity. For instance, in a robotic manipulation task, a physics-grounded model allows a robot to simulate whether a planned grasp will induce excessive deformation or slippage under varying forces before physical execution.

Physical simulation helps understand a real-world system by predicting the next state from the current configuration, leading to broad applications in computer graphics and computational mechanics. The widely used approach is the Material Point Method (MPM)~\cite{Jiang_2015_APIC}, which iteratively computes subsequent steps via a hybrid particle-grid formulation based on material properties and external priors. By combining Lagrangian particles with a background Eulerian grid, MPM effectively handles large deformations and topological changes while retaining the computational convenience and efficiency of regular grids.

Despite its success, most MPM-based reconstruction-driven simulators~\cite{Xie_2024_CVPR_PhysGaussian, Lin_2025_OmniPhysGS, Huang_2024_DreamPhysics, Li_2023_PACNeRF, Cai_2024_GIC, Liu_2024_Physics3D} advance dynamics using explicit time stepping. These methods approximate within-step evolution through start-of-step evaluations, applying a single forward update to the system state. While adequate for sufficiently small intervals, numerical error and instability grow rapidly as the time step increases, particularly when handling stiff materials, strong nonlinearities, or contact-rich dynamics. Consequently, explicit works are strictly constrained by Courant–Friedrichs–Lewy (CFL) conditions~\cite{Courant_1928_CFL}. These stability requirements often necessitate prohibitively small time steps, which not only increase computational overhead but also accumulate discretization errors, thus limiting the practicality of long-horizon physical prediction.

These limitations motivate \emph{within-step} modeling. We propose \textbf{i-PhysGaussian}, which replaces explicit stepping with an \textbf{I}mplicit MPM integrator. 
Concretely, we define a \emph{within-step momentum-balance residual} that measures the mismatch between the end-of-step momentum change and the impulse induced by internal stresses and external loads over the step. We defer the full discretization and its relation to the underlying MPM transfer scheme to \cref{sec:method}.
We then iteratively solve for the end-of-step state so that this residual becomes small, using a Newton-type optimization (with GMRES for the linearized subproblems). 
This implicit formulation substantially relaxes time-step sensitivity and improves physical consistency over a much wider range of step sizes.
While implicit stepping incurs additional per-step optimization, it can reduce the total number of integration steps by enabling larger stable time steps.

Our system comprises two modules: (1) a scene-representation (SR) module built on 3D Gaussian Splatting (3DGS)~\cite{Kerbl_2023_3DGS}, responsible for parsing the initial scene and rendering simulation results; and (2) an implicit physics simulation (i-PS) module built on implicit MPM, responsible for predicting end-of-step states. Given a static 3DGS scene and a configuration file specifying material parameters, simulation domain, and applied loads, the SR module preprocesses the scene and converts it into a set of simulation particles (including surface Gaussians and optionally filled interior particles following PhysGaussian). The i-PS module advances these particles by implicit stepping and returns the updated state for rendering. Repeating this loop yields a dynamic 4D Gaussian sequence, a time-varying set of 3D Gaussians (with positions and optionally other attributes) over discrete time steps.

In summary, our contributions are as follows:
\begin{itemize}
    \item To the best of our knowledge, we are the first to integrate \emph{Newton--GMRES fully implicit} MPM time stepping into a 3DGS-based reconstruction-driven dynamic pipeline for 4D Gaussian sequences.
    \item We introduce a within-step momentum-balance formulation solved by Newton--GMRES, which substantially improves robustness to large time steps and enhances physical consistency compared to explicit baselines.
    \item Open-source implementations of implicit MPM are scarce and predominantly C++; we will release a clean Python implementation to facilitate reproducibility and future research.
\end{itemize}

\subsection{Physics-Based Dynamic Simulation}

With the rise of \emph{explicit} scene representations (\textit{e.g.}, 3DGS), physics-based dynamic simulation has emerged as a promising route for generating temporally coherent dynamics from reconstructed scenes \cite{huang2025towards}. Recent work increasingly integrates differentiable rendering with physical priors to reduce non-physical drift under prescribed material properties and external loads.

Recent methods couple 3DGS with physics solvers for physically plausible 4D Gaussian dynamics. Gaussian Splashing~\cite{Feng_2025_GaussianSplashing} combines 3DGS with Position-Based Dynamics (PBD)~\cite{Muller_2007_PBD, Macklin_2016_XPBD}, while PhysGaussian~\cite{Xie_2024_CVPR_PhysGaussian} integrates 3DGS with MPM~\cite{Sulsky1994MPM, Hu_2018_MLSMPM}. MPM is more suitable for large deformations and contact-rich dynamics and admits effective differentiable implementations. Building on this paradigm, PIDG~\cite{hong2025physics} treats Gaussians as Lagrangian material points and enforces momentum-residual constraints for improved physical consistency. AS-DiffMPM~\cite{vasile2025gaussian} extends differentiable MPM to system identification with arbitrarily shaped colliders, enabling richer object--environment interactions, while GaussianFluent~\cite{huang2026gaussianfluent} further supports mixed materials and brittle fracture via continuum-damage MPM and synthesized volumetric interiors.

A key bottleneck is \emph{time integration}: explicit stepping can become unstable or inaccurate as the time step grows, especially for stiff or highly nonlinear/contact-rich dynamics. Orthogonally, diffusion/video supervision~\cite{Huang_2024_DreamPhysics, Zhang_2024_PhysDreamer, Liu_2024_Physics3D} and short-sequence supervision~\cite{Cai_2024_GIC, Li_2025_CVPR, Li_2025_ICCV} reduce manual configuration by inferring physical parameters, while feed-forward dynamics models (e.g., PhysGM~\cite{lv2025physgm}) amortize reconstruction and prediction for faster 4D synthesis. Interactive systems (e.g., PhysTalk~\cite{collorone2025phystalk}, VR-GS~\cite{jiang2024vr}, GS-Verse~\cite{pechko2025gs}) emphasize real-time control. Even with these advances, robustness under deliberately enlarged time steps remains comparatively underexamined.

\section{Methodology}
\label{sec:method}

\begin{figure*}[t]
  \begin{center}
    {\includegraphics[width=0.90\textwidth]{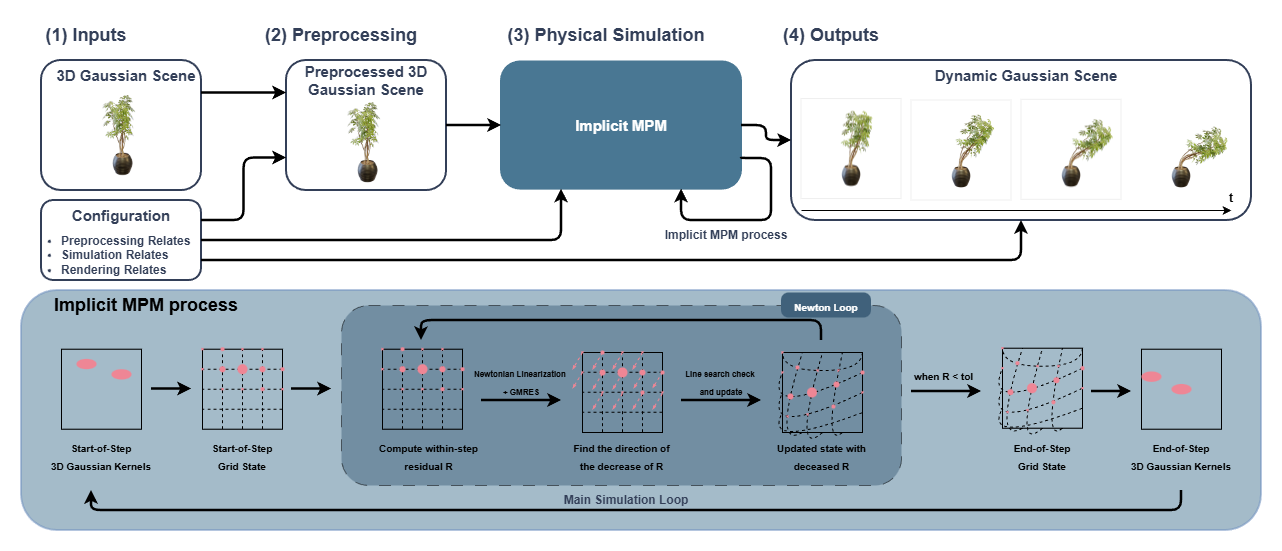}}
    \caption{
      Overview of \textit{i-PhysGaussian}. The pipeline starts with (i) a static 3D Gaussian scene and
      (ii) a configuration file specifying preprocessing, simulation, and rendering options.
      The scene is first preprocessed (axis alignment, pruning near-transparent Gaussians, and optional particle filling),
      then enhanced by an implicit MPM solver under the specified material and boundary conditions.
      The end-of-step state is re-rendered as a Gaussian scene at each time iteratively, which produces a temporally stacked 4D Gaussian representation (4DGS).
    }
    \label{fig:pipeline_overview}
  \end{center}
\end{figure*}

We propose \textit{i-PhysGaussian}, built upon the PhysGaussian framework. As illustrated in Fig.~\ref{fig:pipeline_overview}, the system takes two inputs: (i) a static 3D Gaussian scene and (ii) a configuration file specifying preprocessing, simulation, and rendering options. The pipeline proceeds as follows. First, the input scene is preprocessed (\textit{e.g.}, aligned to the XYZ axes, removing near-transparent Gaussian kernels, and applying particle filling in target regions). Next, the prepared scene, together with the simulation parameters (materials and boundary conditions), is fed to an implicit MPM solver to predict the next state. Each end-of-step state is then re-rendered as a Gaussian scene at the corresponding time instant. Iterating this loop and stacking the per-step scenes along the temporal axis yields a 4D Gaussian representation (4DGS).

The distinction lies in the MPM time integration. Explicit MPM uses quantities evaluated at the beginning of the step (forward Euler) and then transfers back to particles to obtain the end-of-step variables directly. As a result, it is extremely sensitive to step time and prone to collapse in practice. In contrast, we adopt a standard implicit formulation that couples start- and end-of-step variables via a within-step momentum residual, and compute the end-of-step state with Newton–GMRES until the residual meets a prescribed tolerance, thus demonstrating enhanced stability. Details follow in Sec.~\ref{sec:implicit_mpm}.

\subsection{Preliminaries}

We review the standard pipeline and preliminaries for physical simulation under 3D Gaussian scenes.

\paragraph{3D Gaussian Splatting.}

3DGS~\cite{Kerbl_2023_3DGS} represents a static scene as a set of anisotropic 3D Gaussian kernels. Let the complete Gaussian scene be $\{\mathcal{G}_p\}_{p=1}^P$ including $P$ material particles, with each kernelized as $\mathcal{G}_p = (\boldsymbol{\mu}_p, \alpha_p, \boldsymbol{\Sigma}_p, \boldsymbol{\beta}_p)$, where $\boldsymbol{\mu}_p\in\mathbb{R}^3$ is the kernel center, $\alpha_p\in(0,1]$ is the opacity, $\boldsymbol{\Sigma}_p\in\mathbb{R}^{3\times 3}$ is the (positive-definite) covariance matrix, and $\boldsymbol{\beta}_p$ is the spherical-harmonic coefficients for color.

To render a novel view for a given camera, each 3D Gaussian is projected onto the image plane as a 2D elliptical footprint. The color at each pixel can be expressed as
\begin{equation}
\label{equ:gaussian_render}
\!\!\!\mathbf{C}(\mathbf{u})\!=\!\!\!
\sum_{k=1}^{K(\mathbf{u})}
T_k(\mathbf{u})\, a_{\pi(k)}(\mathbf{u})\mathbf{c}_{\pi(k)}(\mathbf{u})
\!+\!
T_{K(\mathbf{u})\!+\!1}(\mathbf{u})\,\mathbf{C}_{\mathrm{bg}},\!\!
\end{equation}
where $\mathbf{u}$ denotes a pixel, $\pi$ is the back-to-front order by Gaussian depth, $\mathbf{c}_{\pi(k)}(\mathbf{u})$ is the RGB Gaussian color, $a_{\pi(k)}(\mathbf{u})\in[0,1]$ is the per-pixel opacity from the projected 2D footprint, and $T_k(\mathbf{u})=\prod_{j<k}\big(1-a_{\pi(j)}(\mathbf{u})\big)$ is the accumulated transmittance. Additionally, the last term is affected by default background Gaussian color $\mathbf{C}_{\mathrm{bg}}$. In our work, we input static 3DGS and interpret each kernel as a simulation particle, which provides seamless compatibility: simulated states can be rendered via Eq.~\eqref{equ:gaussian_render} without additional transformation or conversion. Next, we employ 3DGS under Physical scenarios via internal particle filling.

\paragraph{Internal Particle Filling.}
\label{para:particle_filling}
Since 3DGS places Gaussians primarily on visible surfaces, the reconstructed geometry is hollow. Therefore, it restricts dynamics to a thin shell, causing distortion and collapse under physics settings.

To recover volumetric matter, Xie et al.~\cite{Xie_2024_CVPR_PhysGaussian} proposed an internal particle filling procedure, where the voxelized target region is analyzed by casting a ray along 3 axes, $\pm x$, $\pm y$, $\pm z$. Further, based on the even–odd parity rule, we can decide whether each voxel center aligns inside or outside the shape. For interior points, a filling Gaussian particle is instantiated by employing the appearance parameters $\alpha_{p},\boldsymbol{\beta}_{p}$ from the nearest surface Gaussian. Afterwards, we proceed to the mechanics of continuum movement.

\paragraph{Continuum Mechanics.}
To study the deformation and displacement of objects, we denote an initial object space $\Omega_0$ and a deformed one $\Omega_t$ at time $t$. A material point moves according to a motion function $\mathbf{x}=\phi(\mathbf{X},t)$. The deformation gradient is $\mathbf{F}(\mathbf{X},t)=\nabla_{\mathbf{X}}\phi(\mathbf{X},t)$, which encodes local stretch, rotation, and shear. Further, to ensure physical consistency, there are two constraints, namely mass conservation and momentum conservation. Mass conservation ensures the particles stay consistent during movement without losing or creating new ones, formally,
\begin{equation}
\label{eq:mass_cons}
\int_{B^t} \rho(\mathbf{x},t)\, d\mathbf{x}
\;=\;
\int_{B^0} \rho_0(\mathbf{X})\, d\mathbf{X}\,,
\end{equation}
where $B^0 \subset \Omega_0$ and $B^t=\phi(B^0,t)\subset \Omega_t$ denote the same set of points at the initial and the time-$t$, respectively. Then, momentum conservation keeps the inertia consistent, which is balanced by the sum of internal and external forces.
\begin{equation}
\label{eq:momentum_balance}
\rho(\mathbf{x},t)\,\dot{\mathbf{v}}(\mathbf{x},t)
=
\nabla\!\cdot\!\boldsymbol{\sigma}(\mathbf{x},t)
+
\mathbf{f}^{\mathrm{ext}}(\mathbf{x},t),
\end{equation}
where $\rho\,\dot{\mathbf{v}}$ represents inertia, \textit{e.g.,} momentum rate per unit, $\nabla\!\cdot\!\boldsymbol{\sigma}$ is the internal force density generated by stresses; and $\mathbf{f}^{\mathrm{ext}}$ is the external body-force density, \textit{e.g.}, gravity.

Such a continuum mechanics builds the foundation of motion and deformation. Further, we demonstrate how MPM can explicitly solve the next-step movement prediction.

\paragraph{Explicit Material Point Method.}
\label{para:explicit_mpm}
MPM solves dynamical update by alternating between particles and a background Eulerian grid, which enables efficient and stable computation of movement. It starts from given particle states $\{m_p,\mathbf{x}_p,\mathbf{v}_p,\mathbf{F}_p\}$ at time $t$, and explicitly predict time $t' = t+\Delta t$ via three stages: 
\emph{(i) particle to grid (P2G)} which leverages a shape function $w_{Ip}$ to calculate the influence of particle mass on each grid node $I$, meanwhile creating the grid mass $m_I$ and velocity $\mathbf{v}_I$;
\emph{(ii) Grid update} enables the next-step movement based on physical laws, including velocity $\mathbf{v}_I(t)$ update based on acceleration $\mathbf{a}_I(t)$:
\begin{equation}
\mathbf{v}_I(t') \!=\! \mathbf{v}_I(t) \;+\; \Delta t\, \mathbf{a}_I(t),
\ 
\mathbf{a}_I(t)=\frac{\mathbf{f}^{\mathrm{int}}_I(t)\!+\!\mathbf{f}^{\mathrm{ext}}_I(t)}{m_I(t)};
\end{equation}
Further, \emph{(iii) grid to particle (G2P)} transfers the updated grid velocities back to particles to obtain their positions.
\begin{equation}
\!\mathbf{v}_p(t')\!=\!\sum_I w_{Ip}\,\mathbf{v}_I(t'), 
\ 
\mathbf{x}_p(t')=\mathbf{x}_p(t)+\Delta t\,\mathbf{v}_p(t').
\end{equation}
After new particle dynamics are obtained, it simultaneously updates deformation gradients, \textit{e.g.}, $\mathbf{F}_p$, to ensure consistency for the continuum mechanics.

Such an explicit MPM is simple but can become unstable or inaccurate for large $\Delta t$, especially under stiff materials or contact-rich dynamics. Thus, we are motivated to develop implicit time stepping to enhance MPM.

\subsection{i-PhysGaussian: Implicit Material Point Method}
\label{sec:implicit_mpm}
Building on the above three-stage pipeline, our method starts from a static 3D Gaussian scene based on the material configuration. Further, we align the scene to the eulerian frame with $x$–$y$–$z$ axes and perform internal particle filling in designated regions, which is initialized as $\Omega_{0}, t=0$, with each Gaussian kernel denoted $p$. For particle-to-grid transfers, our shape function $w_{Ip}$ leverages the most common separable cubic B-spline function:
\begin{equation}
\!w_{Ip} \!=\! N\!\left(\frac{x_I-x_p}{h}\right)
             N\!\left(\frac{y_I-y_p}{h}\right)\!
             N\!\left(\frac{z_I-z_p}{h}\right),
\end{equation}
where $(x_p,y_p,z_p)$ is the position of particle $p$, $(x_I,y_I,z_I)$ is the coordinate of grid node $I$, $h$ is the grid spacing, and $N(\cdot)$ denotes the 1D cubic B-spline basis. Further, before G2P, the key contribution lies in our implicit grid update.

\paragraph{Within-step Momentum Residual.}
Instead of accepting the outcome of an explicit update, we implicitly seek an end-of-step state that is consistent with the impulse accumulated \emph{within} a time step. To achieve this, we treat the volume $\Omega_{t}$ as a whole during trial movement to predict its potential outcome. Thus, based on the strong-form momentum balance in Eq.~\eqref{eq:momentum_balance}, we have the weak form for grid node $I$:
\begin{equation}
\small
\label{eq:weak_momentum_node}
\int_{\Omega_t} \!\! \rho N_I\dot{\mathbf v} d\Omega
=
-\!\!\int_{\Omega_t}\!\! \nabla N_I:\boldsymbol{\sigma} d\Omega
+
\int_{\Omega_t}\!\! N_I\mathbf{b}d\Omega
+
\int_{\partial\Omega_t}\!\!\! N_I\mathbf t dS,
\end{equation}
where $N_I$ is the grid shape function, $\mathbf b$ is body-force density, and $\mathbf t$ is prescribed traction.
Following implicit MPM formulations~\cite{GuilkeyWeiss2003ImplicitMPM,SulskyKaul2004ImplicitDynamicsMPM}, we discretize the dynamics on the background grid and use a Newmark family update parameterized by $(\beta,\gamma)$.
Critically, we aim to predict a trial state for the next step. Therefore, we introduce \emph{grid displacement increment} $\Delta\mathbf u_I$ over a step $[t_n,t_{n+1}]$ to understand how the current movement is affected by the mechanics.
Particularly, considering an non-Dirichlet-constrained free node $I\in\mathcal F$, we induce the end-of-step kinematics based on the Newmark relations:
\begin{align}
\label{eq:newmark_kinematics}
&\mathbf a_I^{n+1}(\Delta\mathbf u_I)
\!=\!
\frac{\Delta\mathbf u_I - \Delta t\,\mathbf v_I^{n} - \Delta t^2\left(\frac12-\beta\right)\mathbf a_I^{n}}
{\beta\,\Delta t^2},\\
&\mathbf v_I^{n+1}(\Delta\mathbf u_I)
\!=\!\mathbf v_I^{n}
\!+\!
\Delta t\Big((1\!-\!\gamma)\mathbf a_I^{n}\!+\!\gamma\,\mathbf a_I^{n+1}(\Delta\mathbf u_I)\Big).
\end{align}
To implicitly analyze the end-of-step trial, we quantify the mismatch between current inertia and the sum of internal and external forces via a \emph{within-step momentum residual}:
\begin{equation}
\label{eq:residual_free_nodes}
\mathbf R_I(\Delta\mathbf u)
=
\mathbf f^{\mathrm{ext}}_{I}(\Delta\mathbf u)
+
\mathbf f^{\mathrm{int}}_{I}(\Delta\mathbf u)
-
m_I\,\mathbf a_I^{n+1}(\Delta\mathbf u_I),
\ I\in\mathcal F,
\end{equation}
where $m_I$ is the lumped grid mass for a grid node, and the internal force $\mathbf f^{\mathrm{int}}_{I}(\Delta\mathbf u)$ is accumulated from particles using the trial deformation and stress:
\begin{equation}
\label{eq:internal_force_trial}
\mathbf f^{\mathrm{int}}_{I}(\Delta\mathbf u)\!
=\!
-\!\sum_{p} V_p^{0}\,
\mathbf P_p(\Delta\mathbf u)\,
\nabla_{\mathbf X} N_I(\mathbf x_p),\ 
\mathbf P_p\! =\! \boldsymbol{\tau}_p\,\mathbf F_p^{-T},\!
\end{equation}
where $V_p^0$ is the reference particle volume, $\mathbf F_p$ is the trial deformation gradient, $\boldsymbol{\tau}_p$ is the Kirchhoff stress from the constitutive model, and $\nabla_{\mathbf X}N_I$ denotes the shape-function gradient prescribed by configuration.
Then, we aim to seek the minimal-residual trial that balances the inertial and forces. During this process, we interpolate grid neighbors surrounding the particle to obtain $\mathbf F_p(\Delta\mathbf u)$, further updating the trial \(
\mathbf F_p^{\mathrm{trial}} = (\mathbf I+\Delta t\,\nabla\mathbf v^{n+1})\,\mathbf F_p^{n}
\). As a result, the optimal trial is guaranteed for dynamic equilibrium with superior numerical stability, as shown in our experiments in Sec. \ref{sec:exp}. Under the no-free lunch law, our trial process trades efficiency for physical consistency and temporal robustness, which is justified in Secs.~\ref{sec:exp_phys} and \ref{sec:exp_render}.

\textbf{Dirichlet boundary conditions.}
For Dirichlet-constrained nodes $I\in\mathcal D$ with prescribed movement, we enforce velocity constraints by overwriting their state.
Specifically, let $\mathbf v_I^{\mathrm{tar}}$ be the prescribed velocity. We express the constraint in the same Newmark form by defining a history term $\mathbf v_I^{\mathrm{hist}}$ and a scalar $S=\gamma/(\beta\Delta t)$, both determined by $\mathbf v_I^{n}$ and $\mathbf a_I^{n}$:
\begin{equation}
\label{eq:dirichlet_equation}
S\,\Delta\mathbf u_I \;=\; \mathbf v_I^{\mathrm{tar}} - \mathbf v_I^{\mathrm{hist}},
\quad I\in\mathcal D,
\end{equation}
which directly sets $\Delta\mathbf u_I$ for their end-of-step state.
For evaluation, we only study convergence on free nodes $\mathcal F$, and exclude Dirichlet-constrained nodes to avoid mixing constraint-form and force-form residuals. Next, we carefully explain our Newton-type solver to the implicit MPM.

\subsection{Optimization}
Next, we demonstrate our Newton-GMRES optimization.
\paragraph{Newton outer loop.}
We solve the nonlinear system $\mathbf R(\Delta\mathbf u)=\mathbf 0$ using an inexact Newton method with globalization.
We define the objective
\begin{equation}
\label{eq:phi_objective}
\phi(\Delta\mathbf u)=\frac12\|\mathbf R(\Delta\mathbf u)\|_{\mathcal F}^2,
\end{equation}
where $\|\cdot\|_{\mathcal F}$ is the $\ell_2$ norm accumulated over active free nodes, where $\Delta\mathbf u^{(0)}$ is initialized by a kinematic predictor,
\begin{equation}
\label{eq:du_predictor_method}
\Delta\mathbf u_I^{(0)}=\Delta t\,\mathbf v_I^{n}+\frac12\Delta t^2\,\mathbf a_I^{n}, \quad I\in\mathcal F,
\end{equation}
and then overwrite constrained nodes as in Eq.~\eqref{eq:dirichlet_equation}. At Newton iterate $k$, we linearize the residual around $\Delta\mathbf u^{(k)}$:
\begin{equation}
\label{eq:newton_linearize_du}
\mathbf R(\Delta\mathbf u^{(k)}+\delta\mathbf u)
\approx
\mathbf R(\Delta\mathbf u^{(k)})+\mathbf J(\Delta\mathbf u^{(k)})\,\delta\mathbf u,
\end{equation}
and compute an update direction by approximately solving
\begin{equation}
\label{eq:newton_system_du}
\mathbf J(\Delta\mathbf u^{(k)})\,\delta\mathbf u^{(k)}
=
-\mathbf R(\Delta\mathbf u^{(k)}).
\end{equation}
We employ an Eisenstat--Walker forcing term to adapt the inner linear-solve tolerance, improving efficiency by solving early Newton steps more coarsely and tightening accuracy as the residual decreases.

To globalize the method, we perform a backtracking line search on $\phi$ and update
\begin{equation}
\label{eq:newton_update_line_search}
\Delta\mathbf u^{(k+1)}=\Delta\mathbf u^{(k)}+\alpha^{(k)}\,\delta\mathbf u^{(k)}.
\end{equation}
We use an Armijo decrease test and a weak curvature check based on the directional derivative
\begin{equation}
\label{eq:phi_prime0}
\phi'(0)=\left\langle \mathbf R(\Delta\mathbf u^{(k)}),\; \mathbf J(\Delta\mathbf u^{(k)})\,\delta\mathbf u^{(k)}\right\rangle_{\mathcal F}.
\end{equation}
If $\phi'(0)\ge 0$ (non-descent), we fall back to the steepest-descent direction $\delta\mathbf u=-\mathbf R$ on free nodes.
We terminate when $\|\mathbf R(\Delta\mathbf u)\|_{\mathcal F}$ falls below a prescribed tolerance or when the update stagnates.

\paragraph{GMRES inner loop (right-preconditioned, matrix-free).}
The Newton system~\eqref{eq:newton_system_du} is large and sparse; assembling $\mathbf J$ explicitly is expensive, especially with diverse constitutive models and plasticity.
We therefore use right-preconditioned GMRES and only require Jacobian--vector products (JVPs).

\textbf{Matrix-free Jacobian action.}
Given the current iterate $\Delta\mathbf u$ and a search direction $\mathbf p$, we approximate
\begin{equation}
\label{eq:jvp_fd}
\mathbf J(\Delta\mathbf u)\,\mathbf p
\;\approx\;
\frac{\mathbf R(\Delta\mathbf u+\varepsilon\mathbf p)-\mathbf R(\Delta\mathbf u-\varepsilon\mathbf p)}{2\varepsilon},
\end{equation}
using a centered finite difference.
The perturbation magnitude $\varepsilon$ is chosen adaptively so that $\|\varepsilon\mathbf p\|_{\infty}$ is on the order of $10^{-4}$, which stabilizes the numerical differentiation.
We also project $\mathbf p$ to the free subspace (Dirichlet entries are zeroed), ensuring hard constraints on JVPs.

\textbf{Right preconditioner.}
We employ a diagonal right preconditioner $W$ that combines an inertial term and a stiffness-like diagonal accumulated during residual evaluation:
\begin{equation}
\label{eq:W_def}
W_I
=
\frac{m_I}{\beta\Delta t^2} + K^{\mathrm{diag}}_I,
\quad I\in\mathcal F,
\end{equation}
where $K^{\mathrm{diag}}_I$ includes a material contribution (scaled by $\lambda+2\mu$) and an optional geometric term.
We solve the right-preconditioned system
\begin{equation}
\label{eq:right_precond_system}
\mathbf J(\Delta\mathbf u)\,W^{-1}\mathbf y = \mathbf b,
\quad \mathbf b=-\mathbf R(\Delta\mathbf u),
\end{equation}
by GMRES in the Krylov subspace generated by repeated application of the operator $\mathbf J(\Delta\mathbf u)\,W^{-1}$.
After obtaining $\mathbf y$, the Newton increment is recovered as
\begin{equation}
\label{eq:recover_delta}
\delta\mathbf u = W^{-1}\mathbf y.
\end{equation}

\textbf{GMRES details.}
GMRES builds an orthonormal basis via Arnoldi with (re)orthogonalization, where we use a two-pass Modified Gram--Schmidt for numerical stability, apply Givens rotations to maintain an upper-triangular least-squares system, and stop when the relative residual drops below a tolerance $\eta$ prescribed by the outer Newton loop.
The resulting $\delta\mathbf u$ is returned to the Newton step, followed by line search and constraint overwrite as described above.

% ===== 实验 ===== 
% 在这个章节中，我们将从场景渲染和物理模拟两个方面来衡量我们模型的功能

\section{Experiments}
\label{sec:exp}

In the experiments, we aim to address some core questions: (i) how far $\Delta t$ can be enlarged before runs become BMF-invalid; (ii) how drift evolves with $\Delta t$ together with the BMF gate\footnote{values beyond $k_{\max}$ may be clamp-dominated}; and (iii) whether improved physical simulation comes at the expense of rendering.

\begin{table}[t]
\caption{BMF-gated stability frontier under time-step enlargement.
For each scene and method, we report the largest stable multiplier $k_{\max}$ and \textit{Fail.\%} (percentage of failed multipliers among all tested $k$ values). We use $k_{\max}=0$ to indicate that none of the tested multipliers passes the BMF gate.}
\label{tab:stability_kmax_failrate}
\centering
\small
\begin{sc}
\setlength{\tabcolsep}{0.5pt}
\begin{tabular}{lcccccc}
\toprule
& \multicolumn{2}{c}{\textbf{Ficus}} & \multicolumn{2}{c}{\textbf{Pillow2Sofa}} & \multicolumn{2}{c}{\textbf{Bread}} \\
\cmidrule(lr){2-3}\cmidrule(lr){4-5}\cmidrule(lr){6-7}
Method & $k_{\max}\uparrow$ & {\tiny fail.}\%$\downarrow$ & $k_{\max}\uparrow$ & {\tiny fail.}\%$\downarrow$ & $k_{\max}\uparrow$ & {\tiny fail.}\%$\downarrow$ \\
\midrule
\textit{i-PhysGaussian} & 20 & 0.0  & 20 & 0.0  & 20 & 0.0 \\
PhysGaussian            & 1 & 90.9  & 20 & 0.0  & 8  & 54.5 \\
DreamPhysics            & 2 & 90.9  & 14 & 45.5 & 1  & 90.9 \\
Physics3D               & 0 & 100.0  & 2  & 81.8 & 0  & 100.0 \\
\bottomrule
\end{tabular}
\end{sc}
\end{table}

\subsection{Experiment Setup}
\label{sec:exp_setup}
\paragraph{Datasets.}
We evaluate on two complementary sources that cover both realistic appearance complexity and controlled physical interactions.
\textbf{PhysDreamer Dataset} provides dense reconstructions from real-world captures with complex geometry and high-frequency textures.
We use \emph{Alocasia} to stress thin structures and leaf-level details, and \emph{Hat} to evaluate fabric-like appearance where high-frequency patterns are visually salient.
\textbf{PhysGaussian Dataset} offers synthetic and visually cleaner scenes with controlled dynamics, which are well-suited for isolating simulator behavior under time-step enlargement.
We use \emph{Ficus} (deformation under external loading), \emph{Bread} (large deformation under interaction), and \emph{Pillow2Sofa} (contact-rich motion).
We use \textbf{PhysGaussian Dataset} for physics metrics and \textbf{PhysDreamer Dataset} for rendering metrics.

\paragraph{Baseline.}
We compare against three reconstruction-driven pipelines.
\textbf{PhysGaussian} couples 3DGS with an \emph{explicit} MPM simulator, isolating the effect of explicit vs.\ implicit stepping under the same rendering backbone.
\textbf{DreamPhysics} augments explicit physics with video supervision, testing how guidance interacts with reconstruction-driven dynamics.
\textbf{Physics3D} provides another video-guided baseline included in our unified trace-based evaluation.
In contrast, \textit{i-PhysGaussian} integrates 3DGS with an \emph{implicit} Newton--GMRES MPM solver, enabling stable integration at larger time steps without video supervision. 

In all evaluations, we only transform the position and shape of each 3DGS kernel. Particularly, positions are calculated by the within-step residual solver, while shape is updated via the material constitutive law. The opacity and appearance parameters are kept fixed.

\subsection{Physical Simulation Fidelity}
\label{sec:exp_phys}

Because solvers expose different internal diagnostics, we adopt a unified trace-based protocol.
We export per-particle mass, volume, and trajectories from all methods and compute all metrics from these traces.
We evaluate (i) stability under time-step enlargement, (ii) time-step robustness relative to $1\times\Delta t$ reported together with the BMF gate, and (iii) reference-free physical plausibility at $1\times\Delta t$. 
We provide \textbf{visual comparison videos} in the supplementary material.

\subsubsection{Stability}
\label{sec:exp_stability}

\paragraph{Evaluation Criteria.}
Particles evolve within a bounded simulation domain $[0,\texttt{grid\_lim}]^3$.
Any coordinate that leaves the domain is hard-clamped to the boundary with a small margin $\epsilon=10^{-6}$
(\textit{i.e.}, each component is clipped to $[\epsilon,\texttt{grid\_lim}-\epsilon]$).
We refer to particles that trigger this clamping in a step as \emph{collapsed} particles.

To quantify clamp-dominated degeneracy, we use the \textbf{boundary-hit mass fraction (BMF)},
defined as the fraction of total particle mass belonging to collapsed particles
(see~\ref{App:BMF} for the exact formula).

\textbf{BMF-gated failure.}
Let $T$ be the number of rendered frames.
An iteration is considered \emph{failed} if clamp dominance persists over time, measured by the exceedance ratio $
r \;=\; \frac{1}{T}\sum_{t=1}^{T}\mathbb{I}\!\left[\mathrm{BMF}_t > 0.5\right]$, and $r>0.5$ leads to failure.

For each scene, we sweep time-step multipliers $k\in\{1,2,4,\ldots,20\}$ and apply the above gate to each run.
Based on the pass/fail outcomes, we report two aggregate indicators per method:
(1) the largest multiplier that passes the gate, denoted $k_{\max}$;
and (2) \textit{Fail.\%}, the percentage of failed multipliers among all tested $k$ values.
We also report the overall failure rate aggregated over all settings.

% -------------------- Figure 1: Failure-rate bar chart --------------------
\begin{figure}[t]
    \centering
    \includegraphics[width=\linewidth]{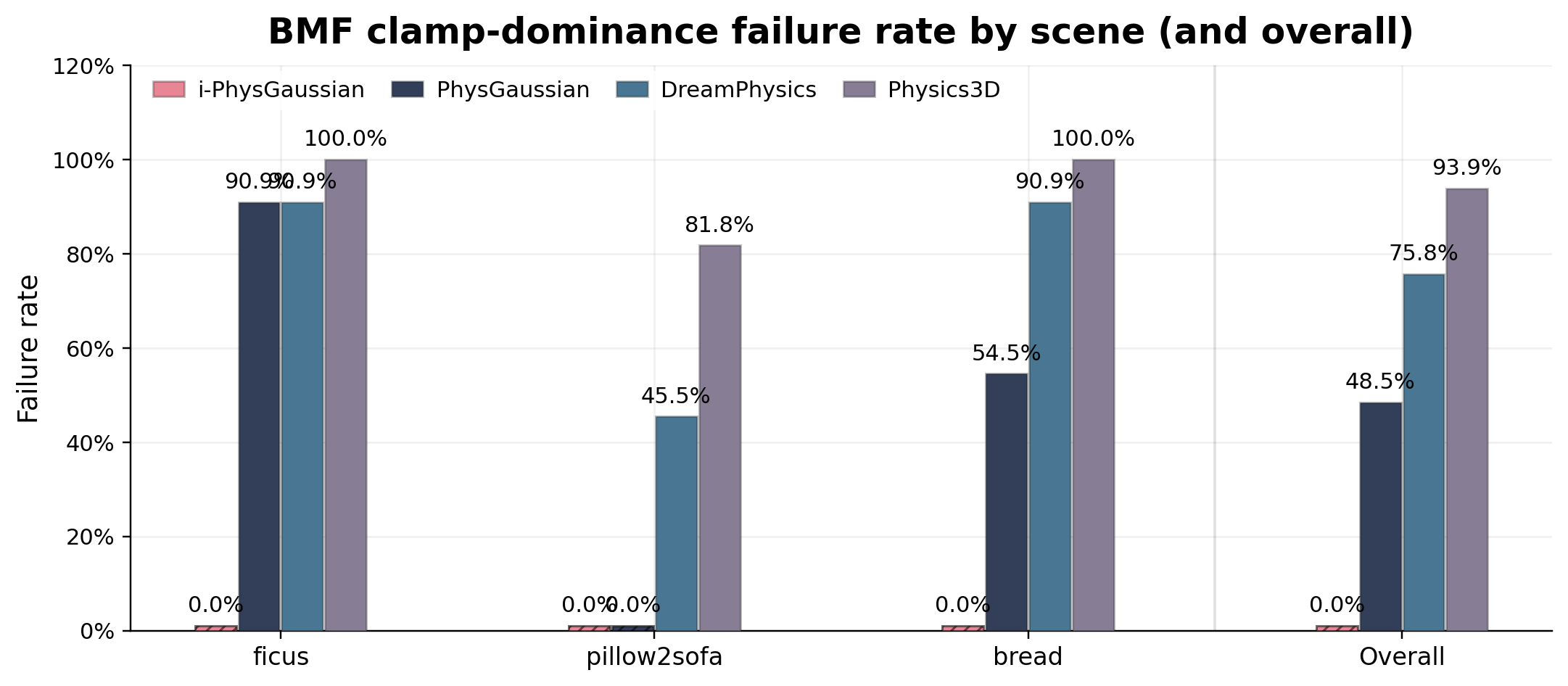}
    \caption{\textbf{BMF-gated failure rate} by scene (and overall). Hatching is used only for visibility; the underlying value remains 0.0\%.}
    \label{fig:bmf_failrate_bar}
\end{figure}

\paragraph{Results and Analysis.}
Tab.~\ref{tab:stability_kmax_failrate} reports the BMF-gated stability frontier ($k_{\max}$) and Fail.\% across time-step multipliers.
\textit{i-PhysGaussian} is uniformly stable, reaching $k_{\max}=20$ with 0.0\% failures on all scenes.
PhysGaussian is stable on Pillow2Sofa ($k_{\max}=20$) but breaks down on Ficus ($k_{\max}=1$) and Bread ($k_{\max}=8$), indicating reduced robustness of explicit stepping under harder dynamics.
DreamPhysics and Physics3D fail much earlier on the challenging scenes, with high failure rates and small $k_{\max}$.

Fig.~\ref{fig:bmf_failrate_bar} visualizes the same trend: \textit{i-PhysGaussian} has 0.0\% failures throughout, while the baselines concentrate failures on Ficus/Bread PhysGaussian or across all scenes (DreamPhysics/Physics3D). For a finer-grained view across all multipliers, we defer the per-$k$ stability heatmaps and additional analyses to Appendix~\ref{App:stability_further}.

\subsubsection{Time-step Robustness}
\label{sec:exp_time_step_robustness}

\paragraph{Evaluation Criteria.}
We quantify how enlarging the time step changes the resulting trajectories relative to the $1\times\Delta t$ reference, and interpret the drift curves jointly with the BMF gate. In particular, results beyond $k_{\max}$ may be dominated by hard clamping and should be treated as diagnostic signals rather than faithful trajectory comparisons.
We report two complementary drift measures:
\textbf{COMD}, which compares the mass-weighted center-of-mass trajectories to the $1\times\Delta t$ reference and reflects \emph{global} motion bias; and
\textbf{mwRMSD}, a mass-weighted particle-level deviation from the same reference that captures \emph{local} deformation and fine-grained motion changes.
To avoid underestimating divergence due to hard boundary clamping, mwRMSD applies a conservative penalty to particles that are clamped in either trajectory.
See Appendix~\ref{App:COMD} and Appendix~\ref{App:mwRMSD} for full definitions.

% ===================== Time-step robustness (single-column, 2x3 clean layout) =====================
\begin{figure}[t]
    \centering

    % ---- Row 1: Ficus ----
    \begin{subfigure}[t]{0.49\columnwidth}
        \centering
        \includegraphics[width=\linewidth]{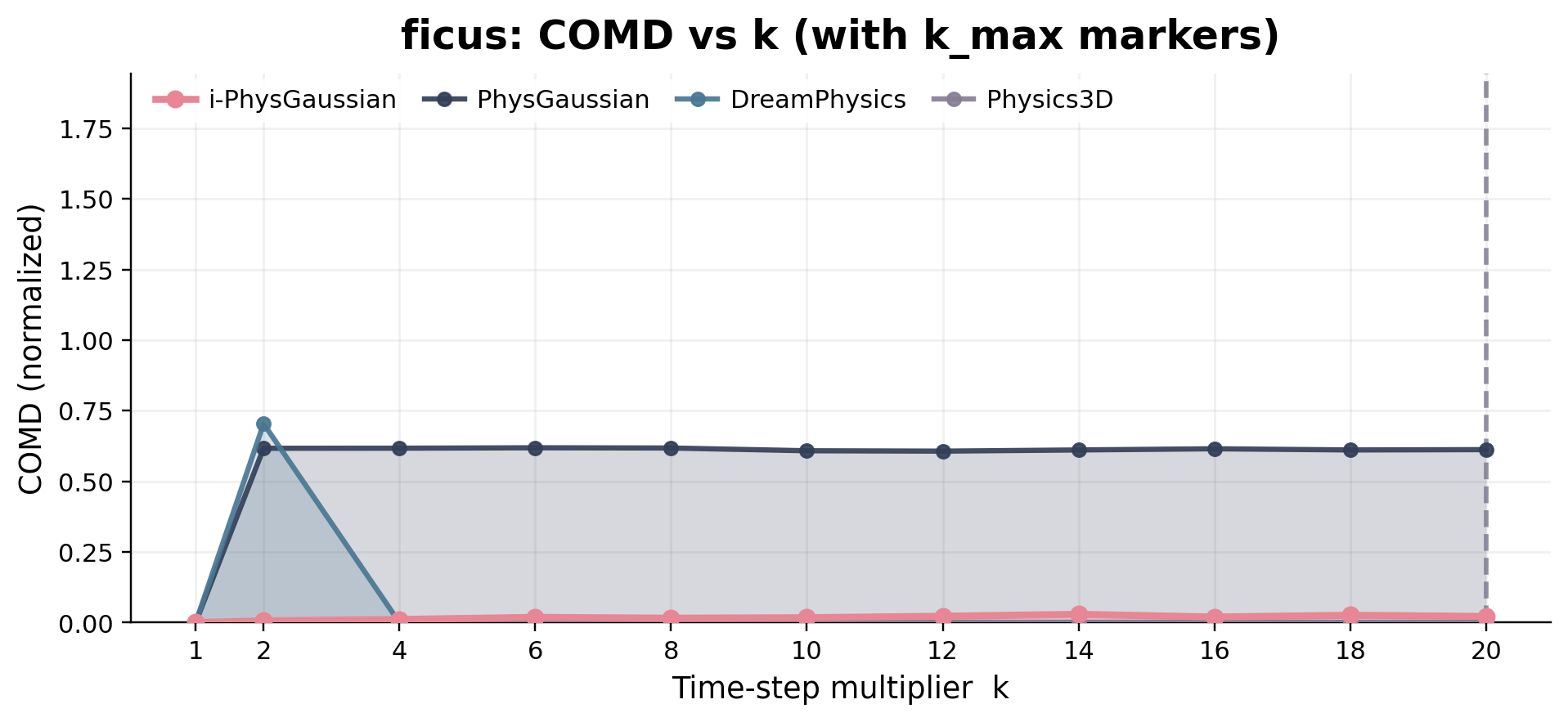}
        \caption{\textbf{Ficus}}
        \label{fig:comd_ficus}
    \end{subfigure}\hfill
    \begin{subfigure}[t]{0.49\columnwidth}
        \centering
        \includegraphics[width=\linewidth]{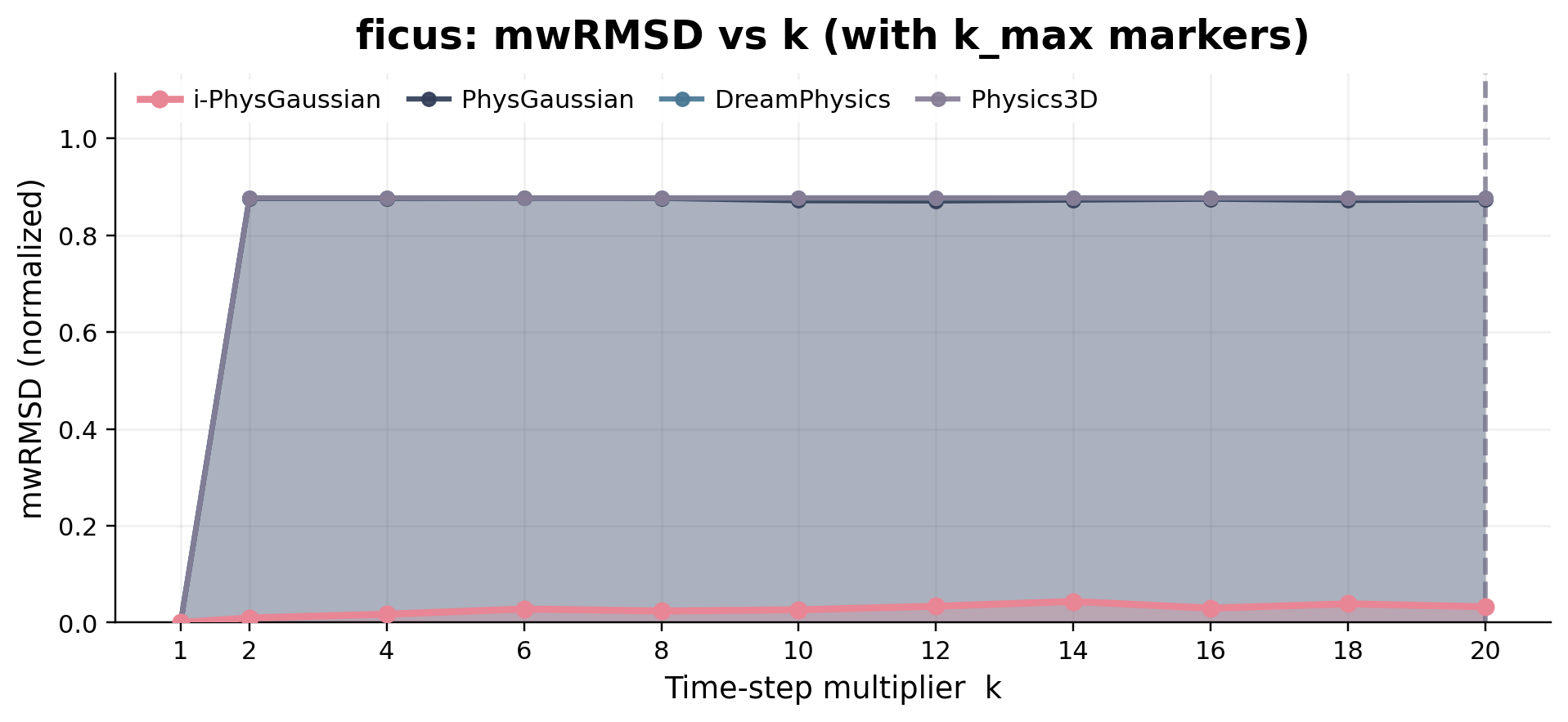}
        \caption{\textbf{Ficus}}
        \label{fig:mwrmsd_ficus_drift}
    \end{subfigure}

    % ---- Row 2: Pillow2Sofa ----
    \begin{subfigure}[t]{0.49\columnwidth}
        \centering
        \includegraphics[width=\linewidth]{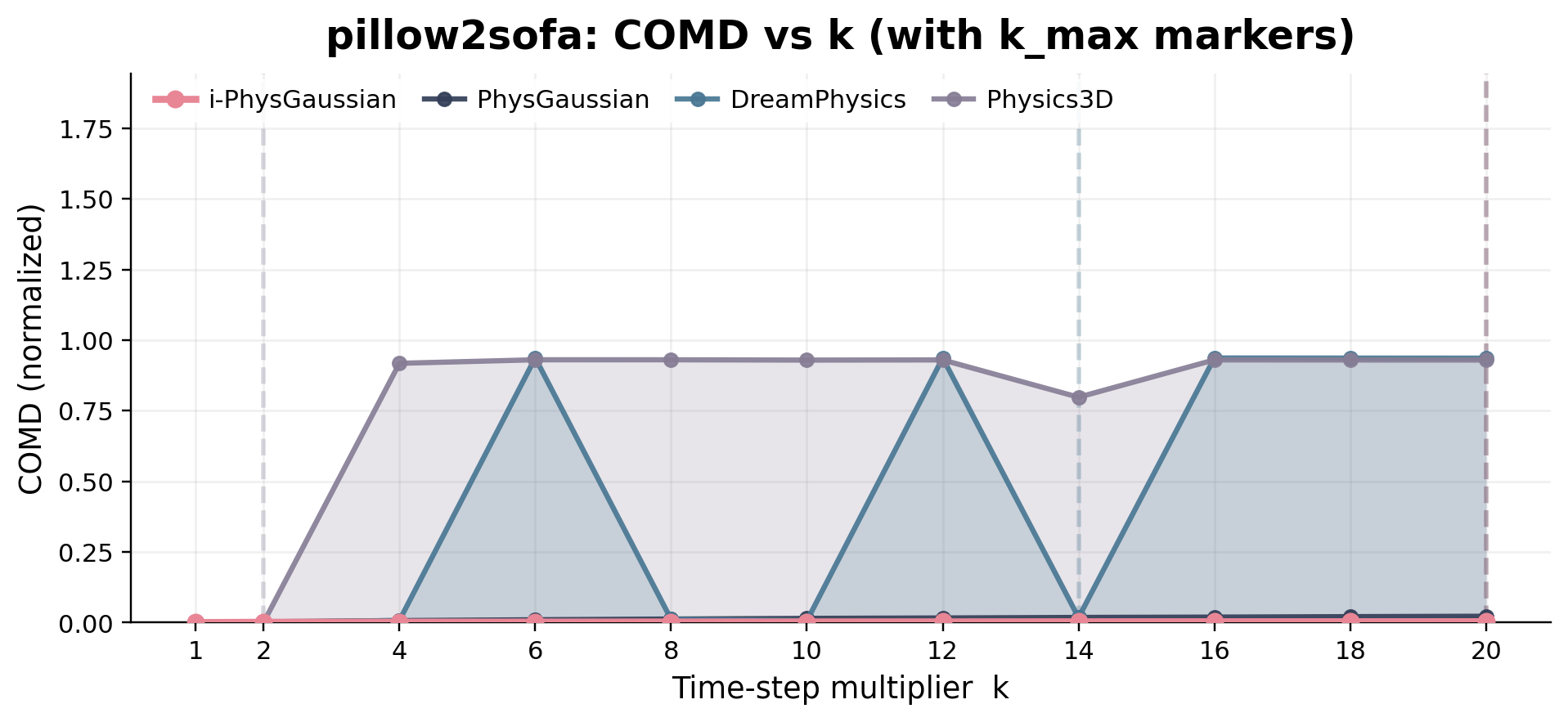}
        \caption{\textbf{Pillow2Sofa}}
        \label{fig:comd_pillow2sofa}
    \end{subfigure}\hfill
    \begin{subfigure}[t]{0.49\columnwidth}
        \centering
        \includegraphics[width=\linewidth]{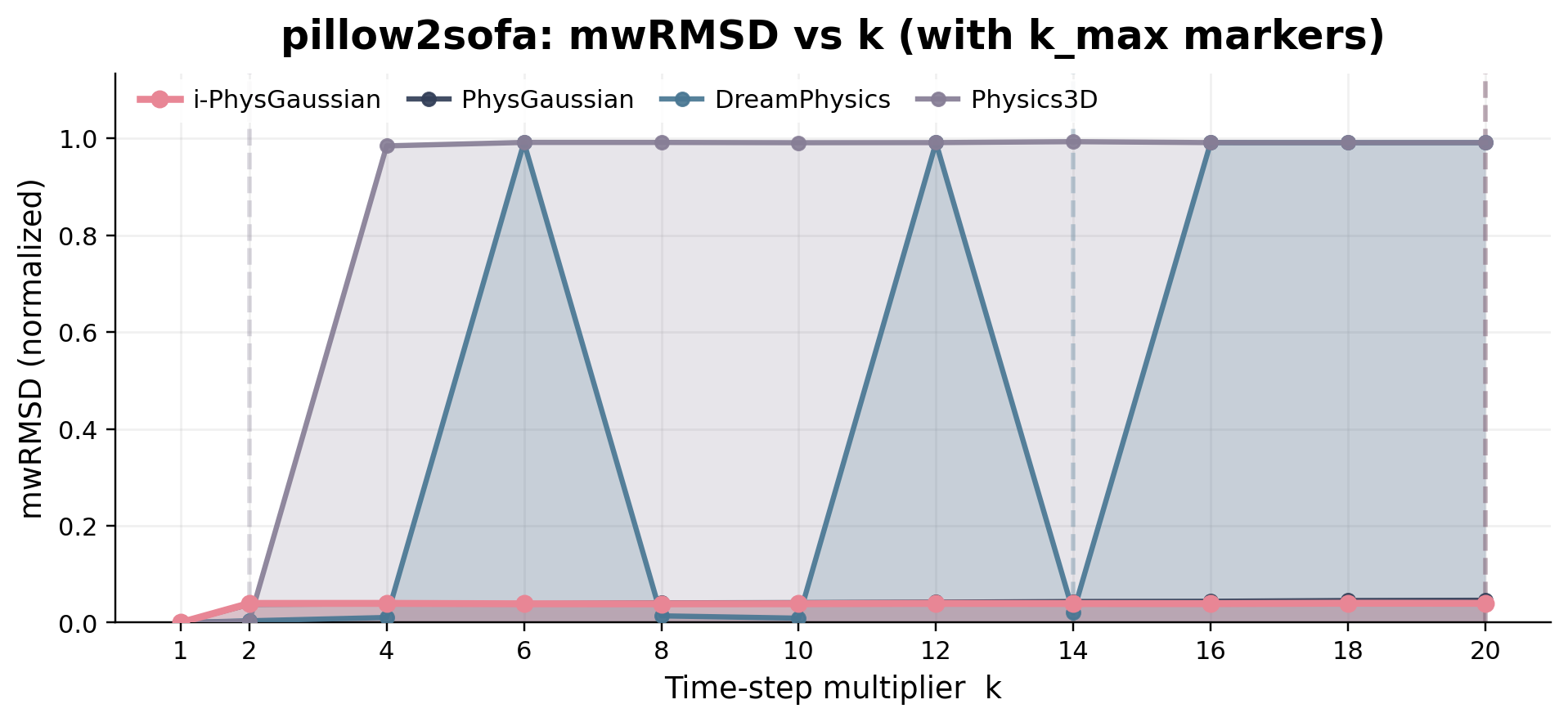}
        \caption{\textbf{Pillow2Sofa}}
        \label{fig:mwrmsd_pillow2sofa_drift}
    \end{subfigure}

    % ---- Row 3: Bread ----
    \begin{subfigure}[t]{0.49\columnwidth}
        \centering
        \includegraphics[width=\linewidth]{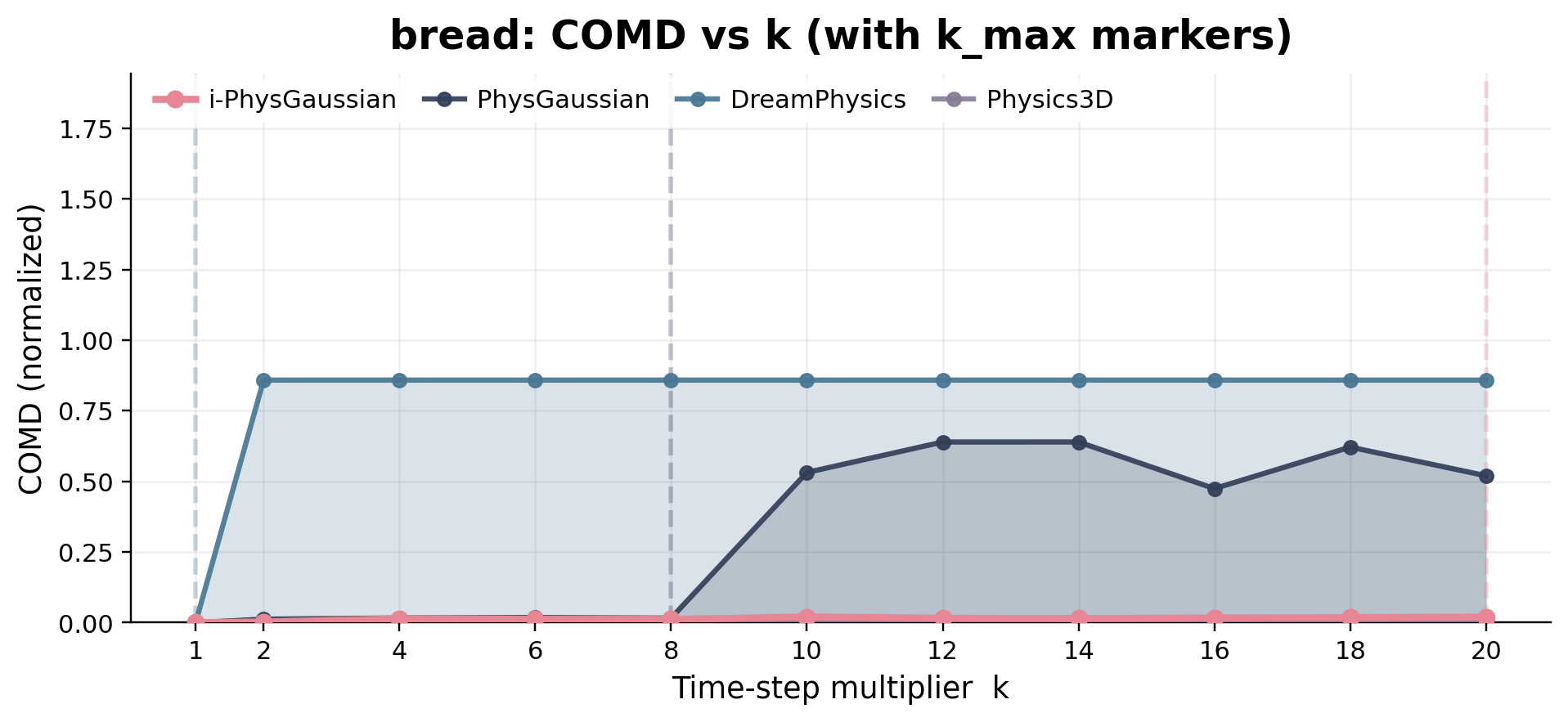}
        \caption{\textbf{Bread}}
        \label{fig:comd_bread}
    \end{subfigure}\hfill
    \begin{subfigure}[t]{0.49\columnwidth}
        \centering
        \includegraphics[width=\linewidth]{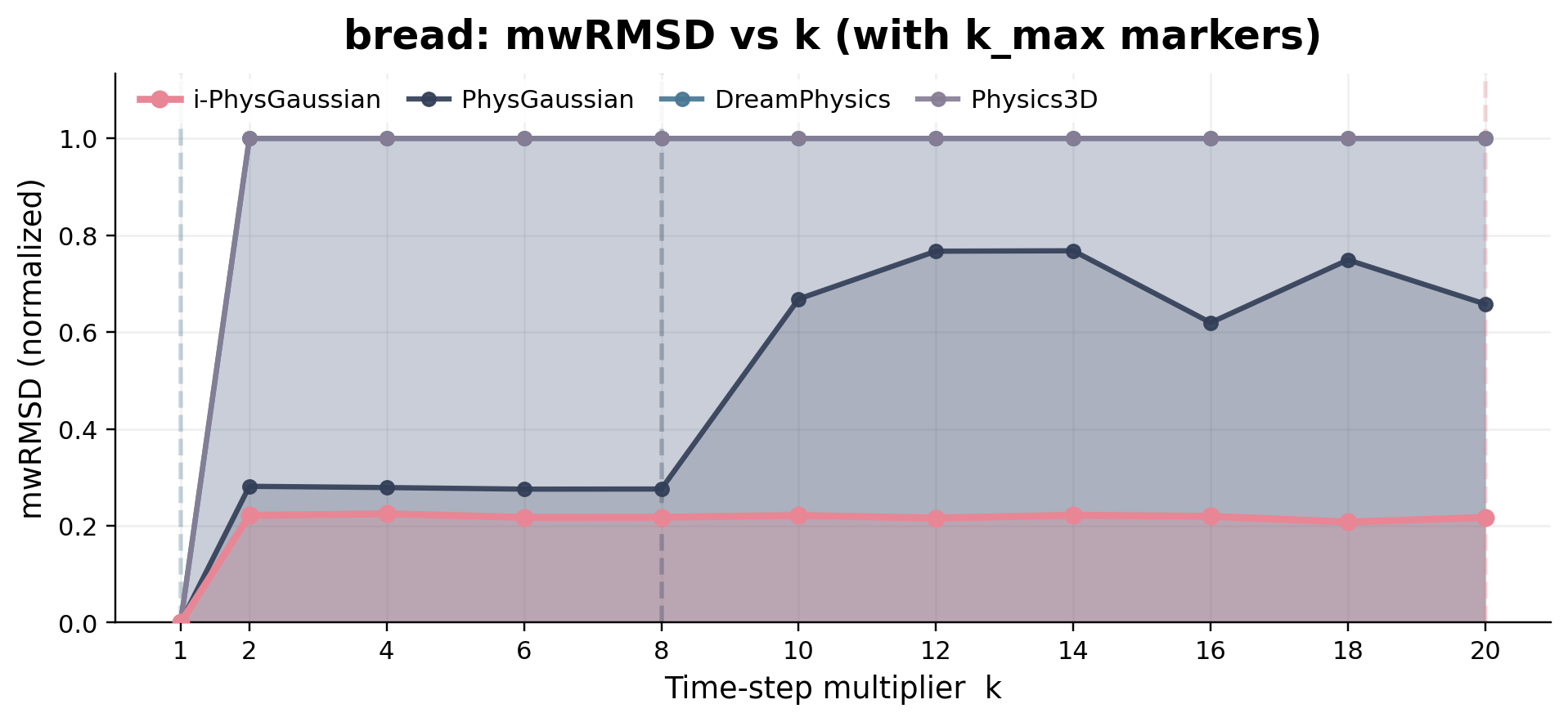}
        \caption{\textbf{Bread}}
        \label{fig:mwrmsd_bread_drift}
    \end{subfigure}
    \caption{\textbf{Time-step robustness with the BMF gate.}
    Left column: COMD$\downarrow$; right column: mwRMSD$\downarrow$.
    Both are plotted versus multiplier $k$ relative to $1\times\Delta t$.
    Dashed lines mark $k_{\max}$; values beyond $k_{\max}$ may be clamp-dominated.}
    \label{fig:dt_robustness_drift}
\end{figure}

\begin{figure}[t]
    \centering

    % ---- Row 1: Ficus ----
    \begin{subfigure}[t]{0.49\columnwidth}
        \centering
        \includegraphics[width=\linewidth]{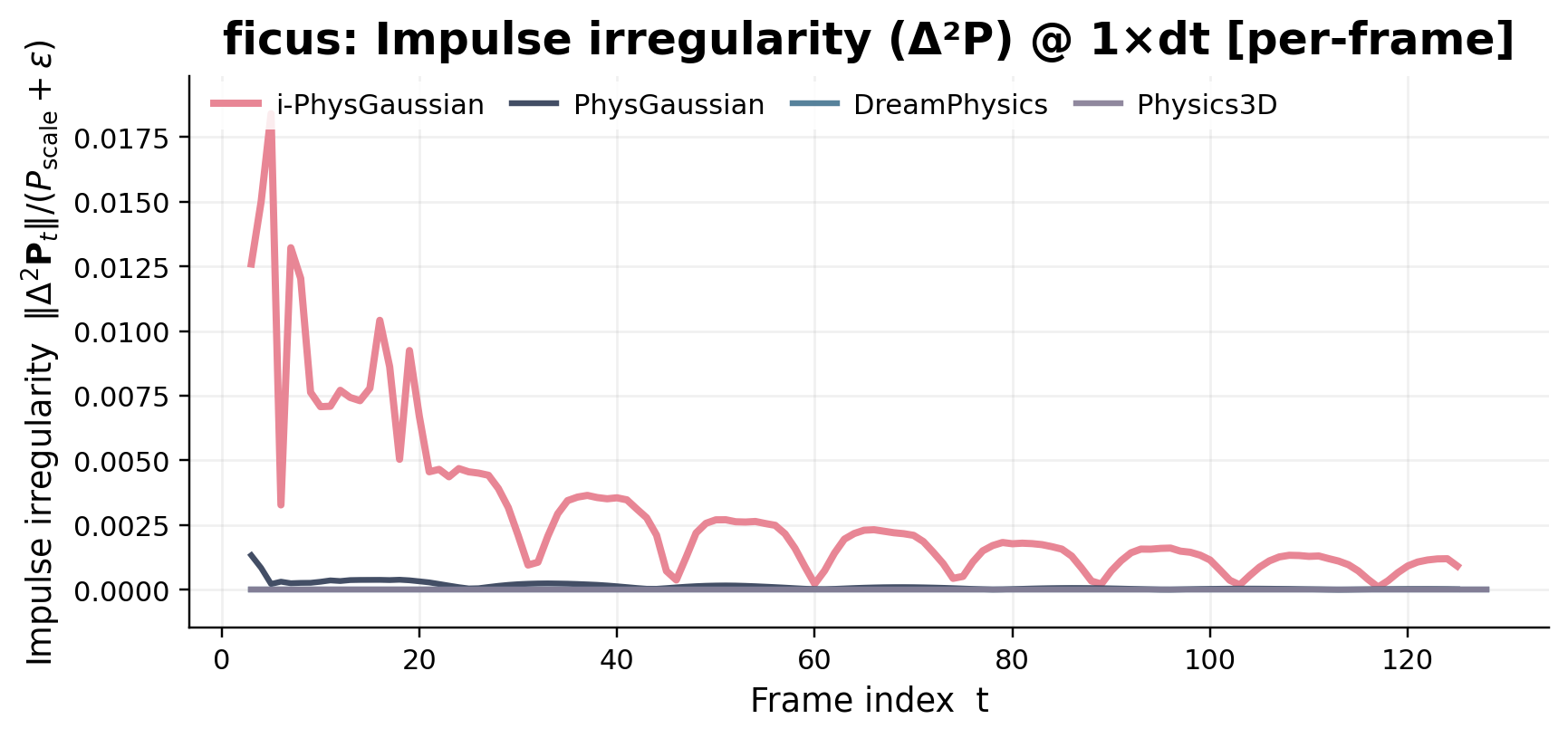}
        \caption{\textbf{Ficus}}
        \label{fig:impirr_ficus}
    \end{subfigure}\hfill
    \begin{subfigure}[t]{0.49\columnwidth}
        \centering
        \includegraphics[width=\linewidth]{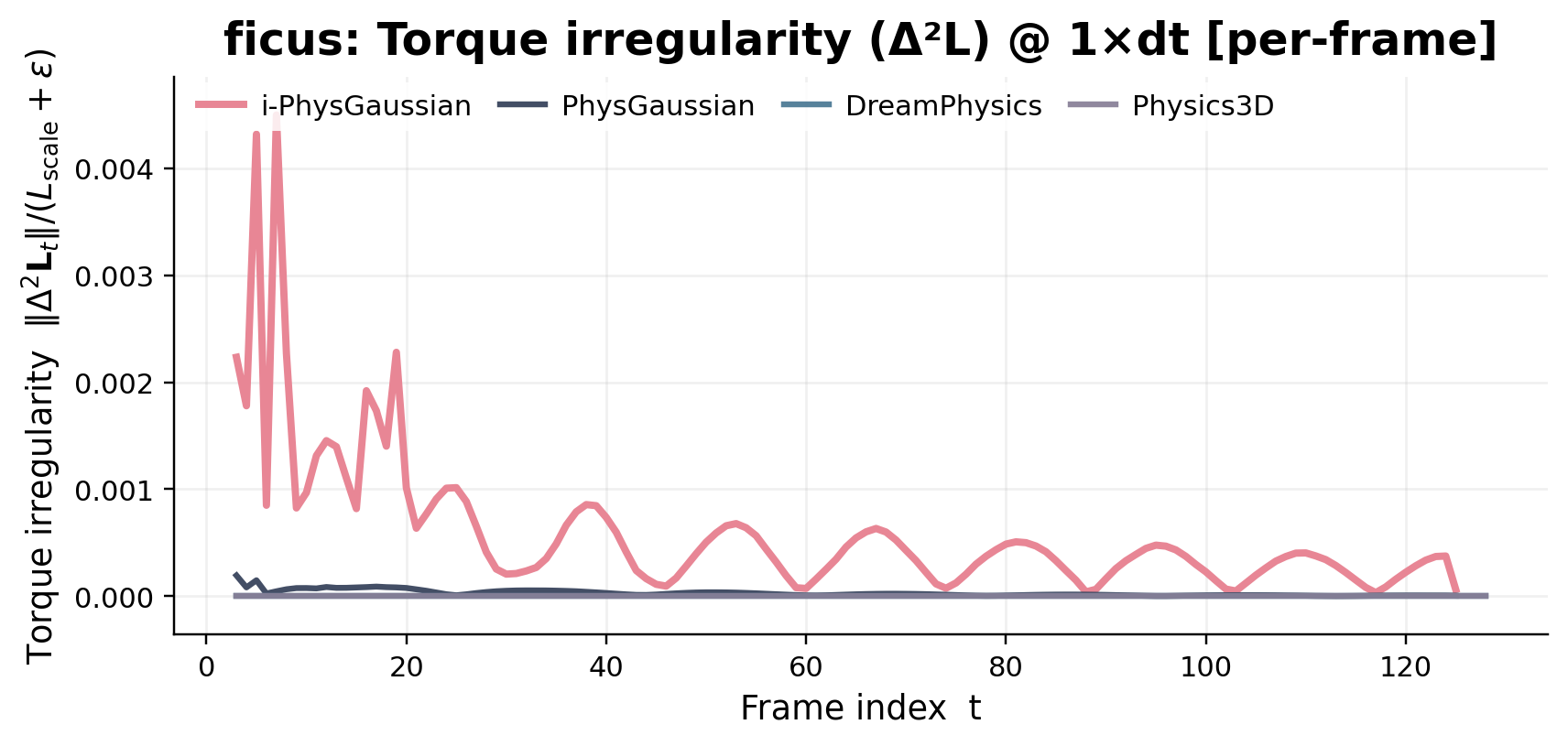}
        \caption{\textbf{Ficus}}
        \label{fig:torirr_ficus}
    \end{subfigure}

    % ---- Row 2: Pillow2Sofa ----
    \begin{subfigure}[t]{0.49\columnwidth}
        \centering
        \includegraphics[width=\linewidth]{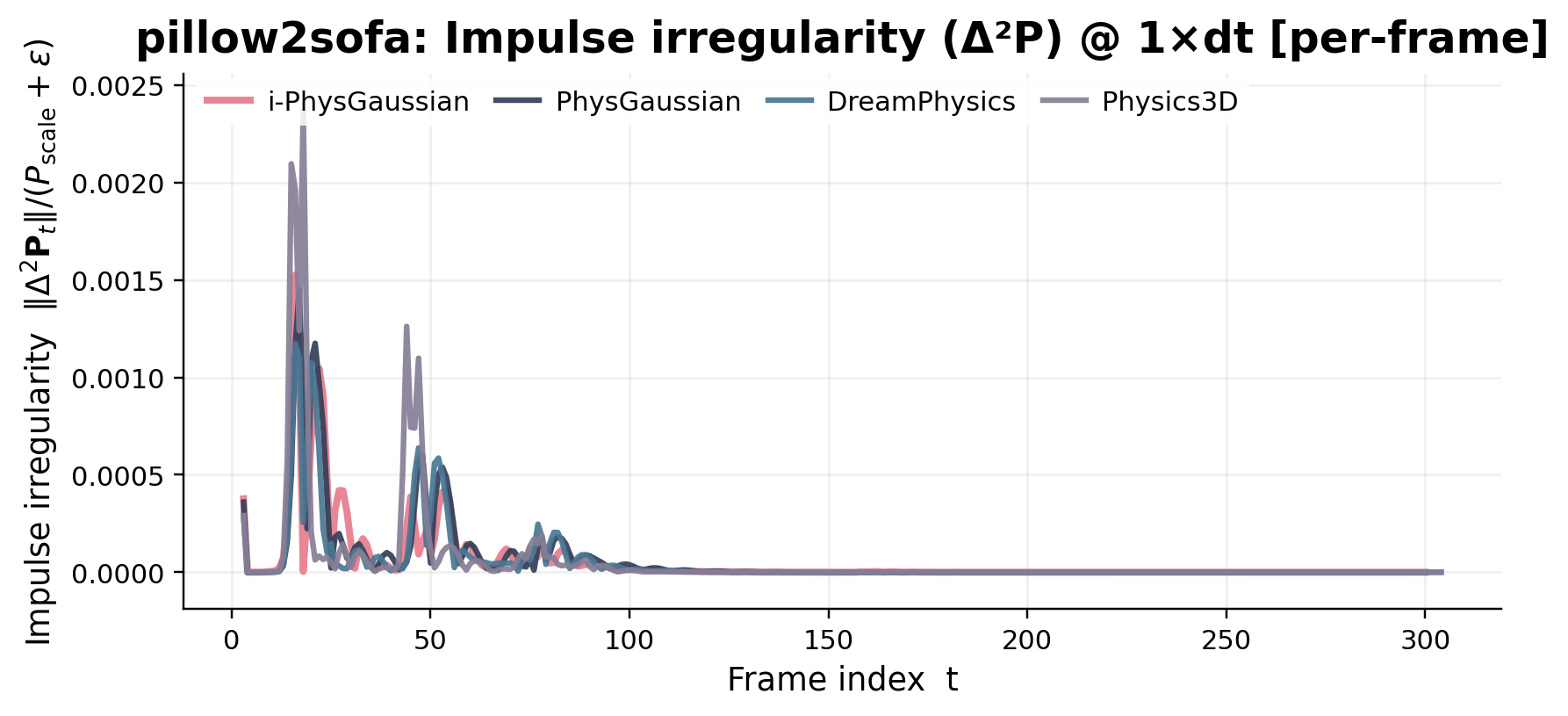}
        \caption{\textbf{Pillow2Sofa}}
        \label{fig:impirr_pillow2sofa}
    \end{subfigure}\hfill
    \begin{subfigure}[t]{0.49\columnwidth}
        \centering
        \includegraphics[width=\linewidth]{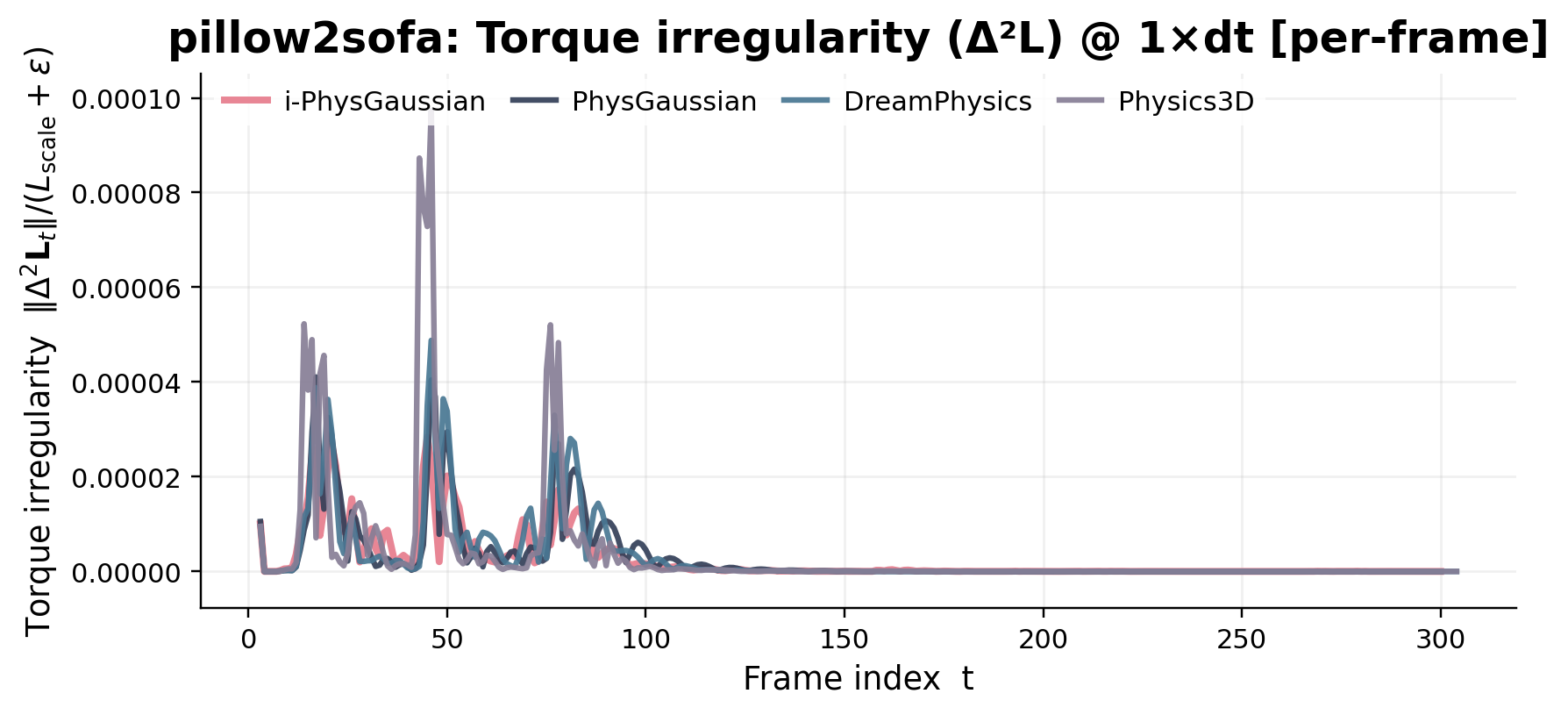}
        \caption{\textbf{Pillow2Sofa}}
        \label{fig:torirr_pillow2sofa}
    \end{subfigure}

    % ---- Row 3: Bread ----
    \begin{subfigure}[t]{0.49\columnwidth}
        \centering
        \includegraphics[width=\linewidth]{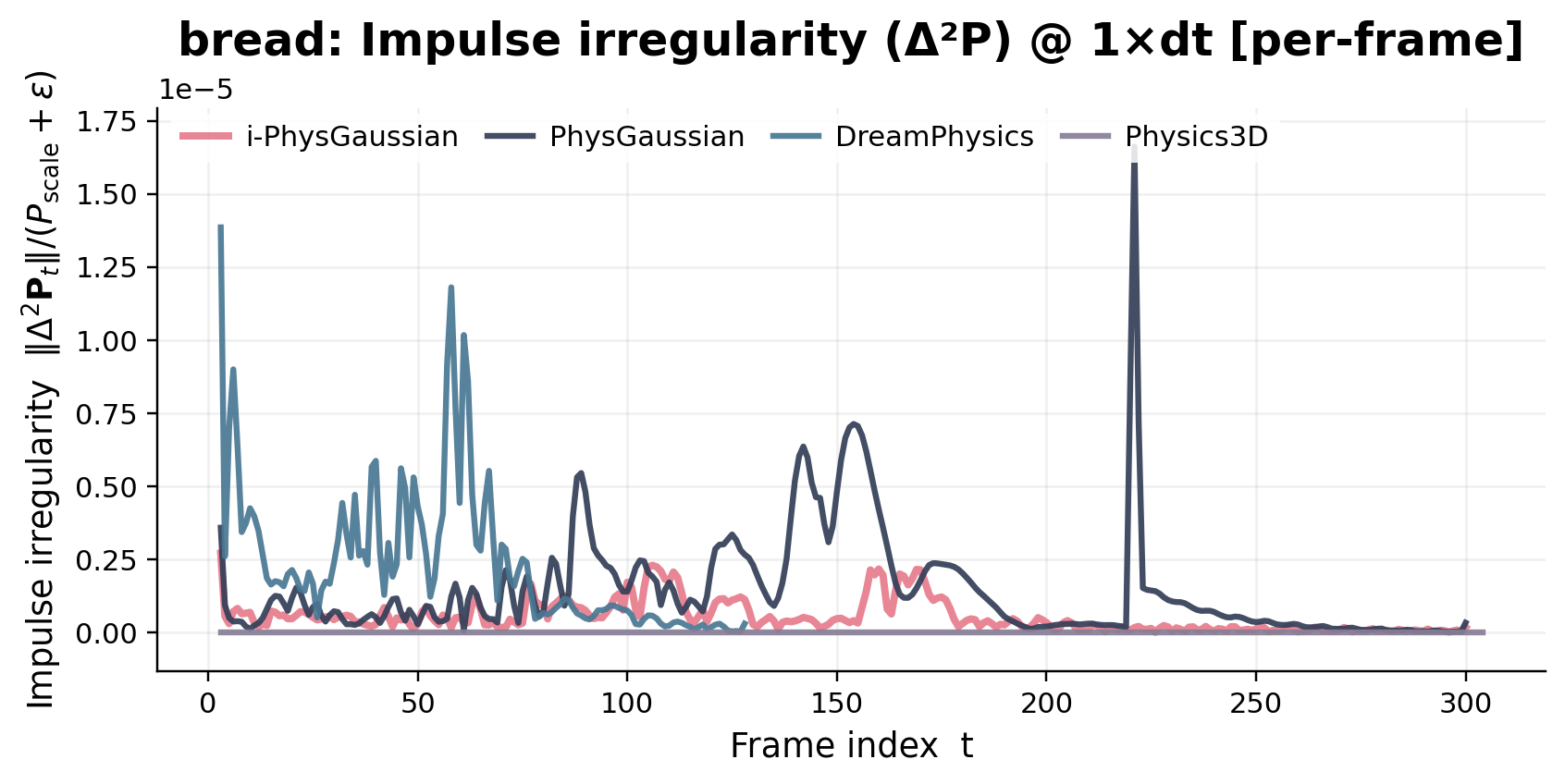}
        \caption{\textbf{Bread}}
        \label{fig:impirr_bread}
    \end{subfigure}\hfill
    \begin{subfigure}[t]{0.49\columnwidth}
        \centering
        \includegraphics[width=\linewidth]{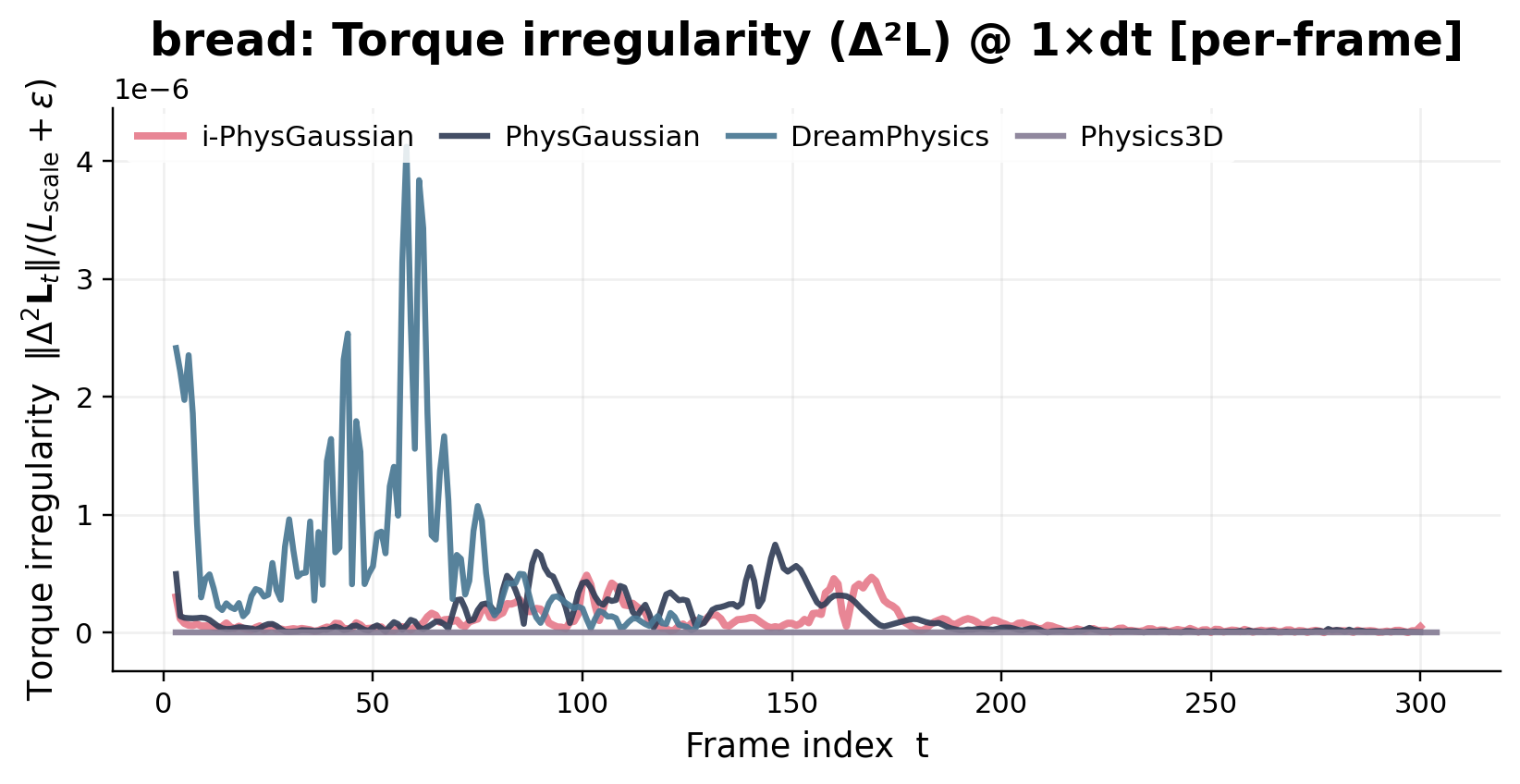}
        \caption{\textbf{Bread}}
        \label{fig:torirr_bread}
    \end{subfigure}
    \caption{\textbf{Reference-free impulse/torque irregularity at $1\times\Delta t$ (per-frame).}
    Left column: impulse irregularity$\downarrow$; right column: torque irregularity$\downarrow$.
    Lower values indicate smoother evolution of net impulse/torque (\textit{i.e.} fewer bursty non-physical jitters).}
    \label{fig:irregularity_1xdt}
\end{figure}

\begin{figure*}[t]
    \centering
    \begin{subfigure}[t]{0.32\textwidth}
        \centering
        \includegraphics[width=\linewidth]{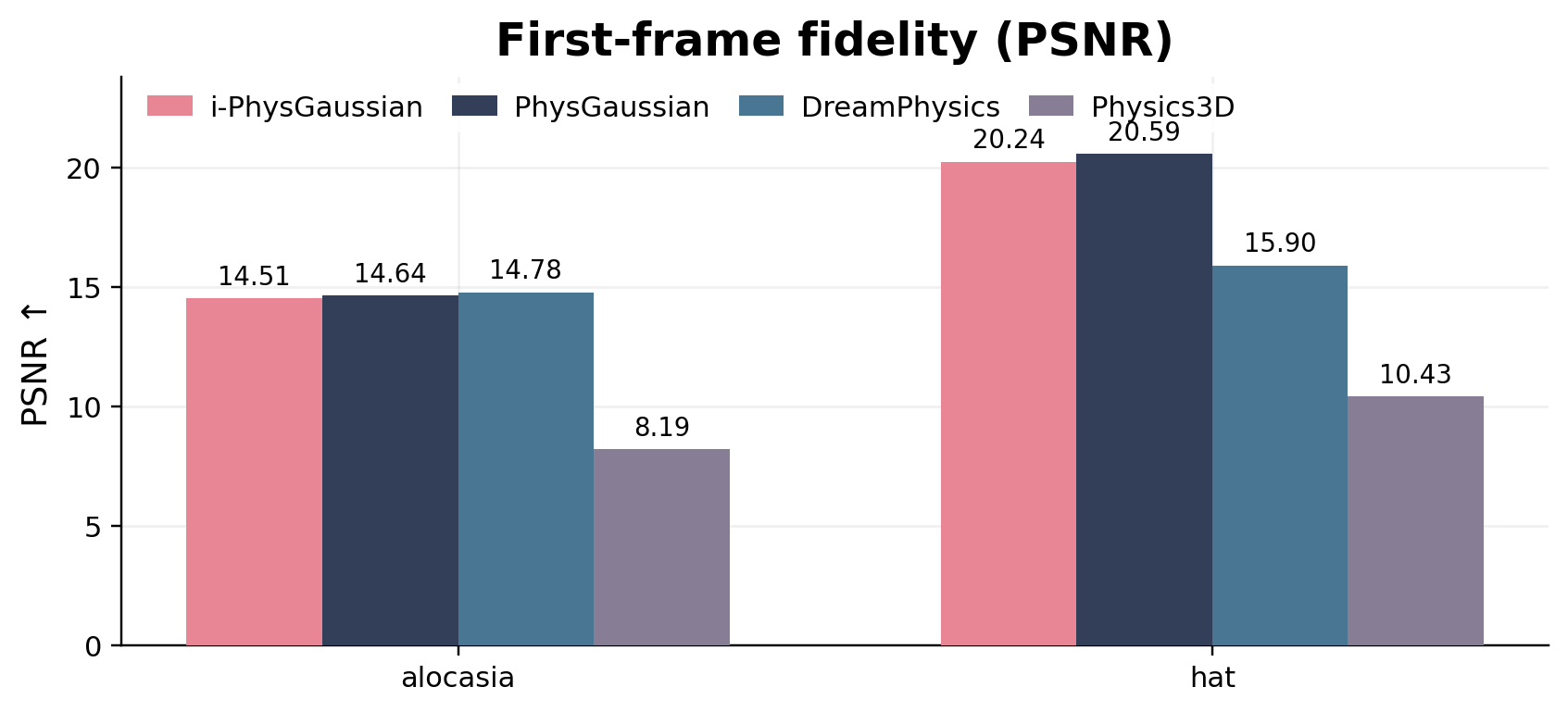}
        \caption{\textbf{PSNR} ($\uparrow$)}
        \label{fig:firstframe_bar_psnr}
    \end{subfigure}\hfill
    \begin{subfigure}[t]{0.32\textwidth}
        \centering
        \includegraphics[width=\linewidth]{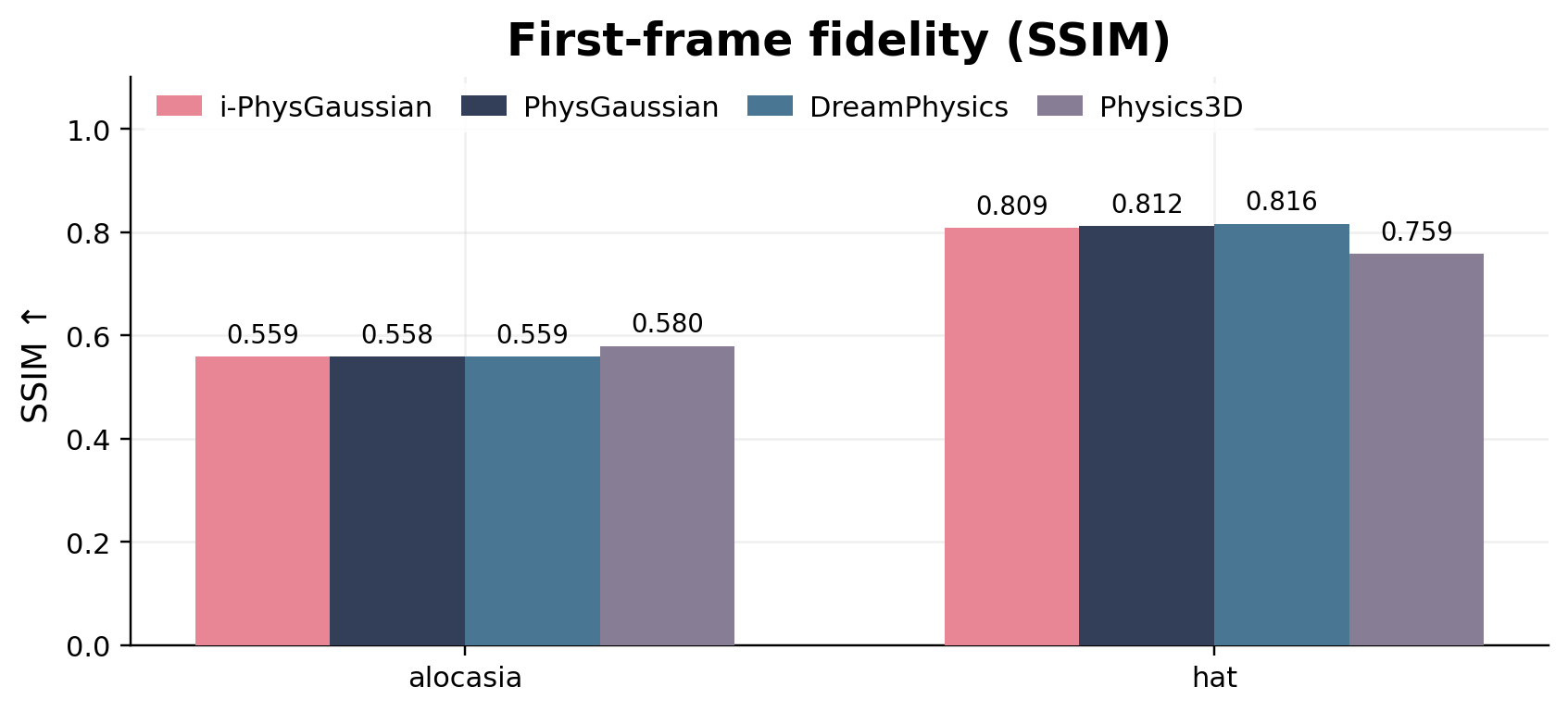}
        \caption{\textbf{SSIM} ($\uparrow$)}
        \label{fig:firstframe_bar_ssim}
    \end{subfigure}\hfill
    \begin{subfigure}[t]{0.32\textwidth}
        \centering
        \includegraphics[width=\linewidth]{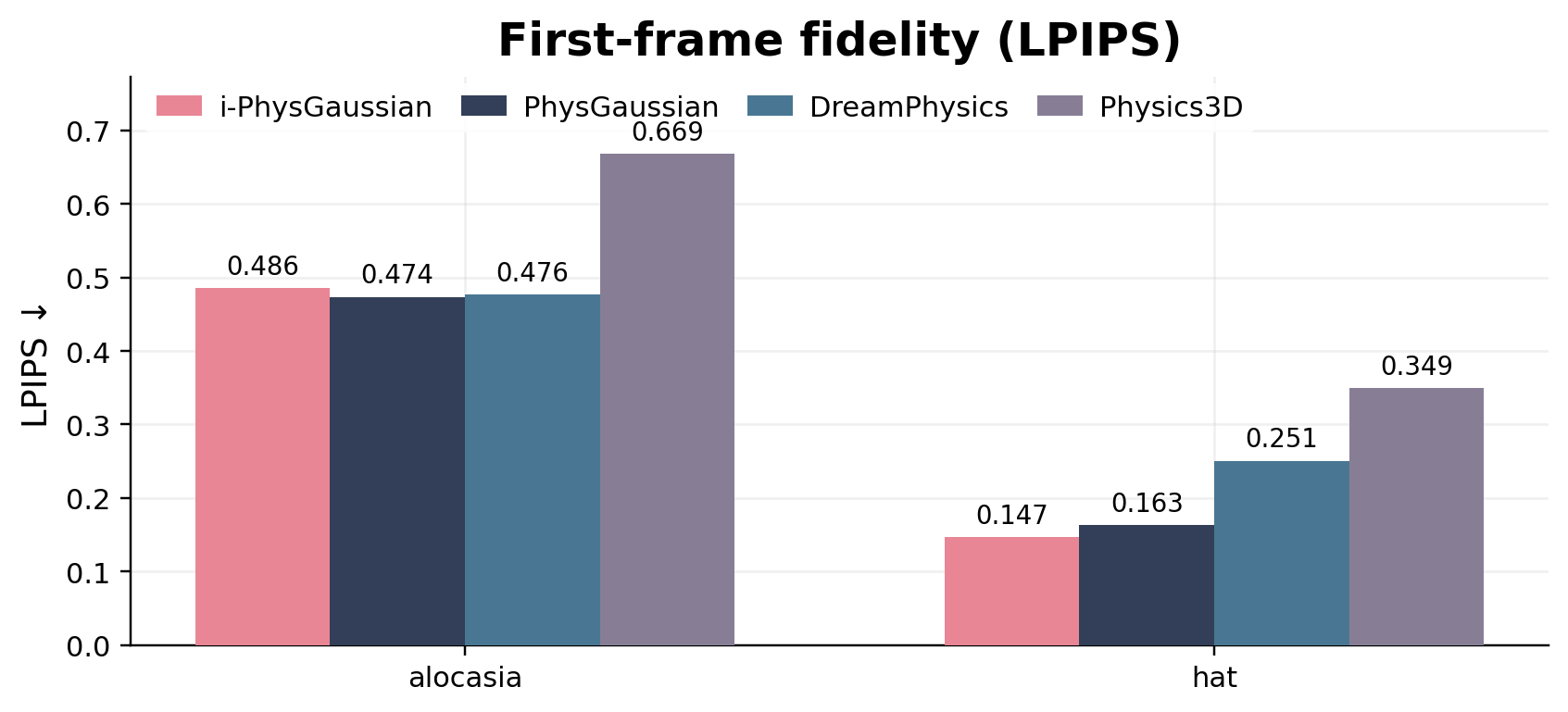}
        \caption{\textbf{LPIPS} ($\downarrow$)}
        \label{fig:firstframe_bar_lpips}
    \end{subfigure}
    \caption{\textbf{Static rendering fidelity at $1\times\Delta t$ (first frame).}
    We compare the first rendered frame against the paired ground-truth image and report PSNR/SSIM/LPIPS
    (higher is better for PSNR/SSIM; lower is better for LPIPS).}
    \label{fig:firstframe_fidelity_bars}
\end{figure*}

\paragraph{Results and Analysis.}
\label{sec:exp_dt_robust}

Stability does not necessarily imply trajectory correctness. Since particle-level ground truth is unavailable, we treat the trajectory $1\times\Delta t$ as a reference and measure how trajectories drift as the time step is enlarged.
To avoid subtle drift caused by boundary projection, mwRMSD applies a conservative penalty to particles clamped in either trajectory, as shown in Appendix~\ref{App:mwRMSD}.

Fig.~\ref{fig:dt_robustness_drift} reports drift versus the multiplier $k$ (left: COMD; right: mwRMSD), with dashed lines indicating each method's stability frontier $k_{\max}$.
Across all three scenes, \textit{i-PhysGaussian} consistently exhibits the lowest drift and the slowest growth with $k$, indicating improved robustness at both global (center-of-mass) and particle levels.
PhysGaussian remains competitive on Pillow2Sofa but deviates noticeably on Bread as $k$ increases, while DreamPhysics and Physics3D show larger drift and stronger variability on the harder scenes.
For a compact cross-$k$ summary, we additionally report normalized AUC scores in Appendix~\ref{App:AUC}.

\subsubsection{Physical Plausibility}
\label{sec:exp_physical_plausibility}

\paragraph{Evaluation Criteria.}
Stability and trajectory-level consistency do not necessarily imply \emph{physical plausibility}.
When particle-level ground truth is unavailable, our robustness metrics treat the $1\times\Delta t$ trajectory as a proxy reference; however, this reference itself may deviate from basic conservation principles due to numerical errors~\cite{wang2024noisegpt,huang2021universal}, hard clamping, or implementation details.
To reduce reliance on a potentially imperfect reference, we conduct a \textbf{reference-free} physical-consistency diagnosis at \textbf{$1\times\Delta t$} for each scene and method.

We report three trace-derived diagnostics from two complementary aspects---\textbf{mass} and \textbf{momentum}:
(1) \textbf{mass drift (MassDrift)}, which checks whether the total mass changes over time;
(2) \textbf{impulse irregularity} ($\Delta^2\mathbf{P}$), which captures high-frequency irregular variations of the total linear momentum; and
(3) \textbf{torque irregularity} ($\Delta^2\mathbf{L}$), which captures high-frequency irregular variations of the total angular momentum.
All quantities are computed directly from exported traces and require no external-force decomposition.
Full details are shown in Appendices~\ref{App:MassDrift} and ~\ref{App:PhysPlausMom}.

\paragraph{Results and Analyses.}
We confirm that mass conservation for all methods and scenes at $1\!\times\!\Delta t$, as shown in Appendix~\ref{App:MassSanity}. Here, we focus on momentum-based diagnostics.
Fig.~\ref{fig:irregularity_1xdt} reports \textbf{reference-free} impulse and torque irregularity, capturing bursty high-frequency jitters in the evolution of net linear/angular momentum. Across scenes, \textit{i-PhysGaussian} exhibits smoother curves with fewer spikes, indicating a continuous evolution of net impulse/torque.

Importantly, near-zero irregularity does not necessarily imply better physical behavior.
In \textbf{Ficus} and \textbf{Bread}, some baselines produce almost flat curves close to zero; the trajectory overlays in Fig.~\ref{fig:key_traj_overlays} suggest this is more consistent with \textbf{kinematic degeneration} toward an almost static trajectory, where motion amplitude is strongly suppressed, and thus momentum variations collapse.
In \textbf{Pillow2Sofa}, all methods show aligned peaks that coincide with major contact/collision moments, supporting that these diagnostics capture physically interpretable events.
Additional discussion and details are provided in Appendix~\ref{App:PhysPlausPerScene}.

\subsection{Dynamic Rendering Quality}
\label{sec:exp_render}

We further examine the \textbf{rendering-side} behavior of physics-driven pipelines to verify that introducing an \emph{implicit} MPM solver does not compromise visual quality.
Since rendering differences are most evident on \textbf{real-captured, texture-rich scenes}, we adopt a complementary protocol: (i) first-frame fidelity against paired ground truth, and (ii) reference-free \textbf{exposure stability} over the rendered sequence.

\subsubsection{Static Rendering Fidelity}
\label{sec:exp_render_static}

\paragraph{Evaluation Criteria.}
We evaluate the first rendered frame ($t{=}1$) with the paired ground-truth image of each model and report \textbf{PSNR/SSIM/LPIPS} to capture pixel-wise structure and perceptual fidelity.

% ===================== First-frame rendering fidelity (bar charts) =====================

\paragraph{Results and Analysis.}
Fig.~\ref{fig:firstframe_fidelity_bars} summarizes first-frame fidelity on two real-captured scenes.
Overall, \textit{i-PhysGaussian} matches PhysGaussian across PSNR/SSIM/LPIPS, indicating no systematic rendering degradation from using an implicit solver.
DreamPhysics is competitive on \textbf{Alocasia} but drops more on \textbf{Hat}, suggesting scene-dependent sensitivity.
Physics3D performs noticeably worse in PSNR/LPIPS, implying larger pixel-level and perceptual deviations, while SSIM differences are generally smaller across methods.

\subsubsection{Dynamic Exposure Stability}
\label{sec:exp_render_exposure}

\paragraph{Evaluation Criteria.}
Since real-captured scenes typically lack frame-aligned ground-truth videos, we adopt a \textbf{reference-free} exposure diagnostic.
We measure the per-frame \textbf{saturation ratio} (Sat.\%), i.e., the fraction of near-saturated (blown-out) pixels, and summarize the sequence by its mean/std/range (max--min).

\begin{table}[t]
\caption{\textbf{Over-exposure diagnostic at $1\times\Delta t$.}
We report saturation ratio statistics (mean/std/range), where SatRatio$_t$ is the percentage of pixels whose luminance satisfies $Y_t(\mathbf{p})\ge 0.98$.
Lower values indicate fewer over-exposed regions.}
\label{tab:overexposure_sat_ratio}
\centering
\small
\begin{sc}
\setlength{\tabcolsep}{1.2pt}
\renewcommand{\arraystretch}{1.05}
\begin{tabular}{lccc}
\toprule
Method & Sat.\% mean$\downarrow$ & Sat.\% std$\downarrow$ & Sat.\% range$\downarrow$ \\
\midrule
\multicolumn{4}{c}{\textbf{Alocasia}} \\
i-PhysGaussian & 0.15  & 0.0105 & 0.0805 \\
PhysGaussian   & 0.15  & 0.0005 & 0.0046 \\
DreamPhysics   & 0.17  & 0.0003 & 0.0017 \\
Physics3D      & 98.14 & 0.0000 & 0.0000 \\
\midrule
\multicolumn{4}{c}{\textbf{Hat}} \\
i-PhysGaussian & 0.05  & 0.0005 & 0.0053 \\
PhysGaussian   & 0.05  & 0.0017 & 0.0218 \\
DreamPhysics   & 0.43  & 0.0000 & 0.0000 \\
Physics3D      & 92.93 & 0.0000 & 0.0000 \\
\bottomrule
\end{tabular}
\end{sc}
\end{table}

\paragraph{Results and Analysis.}
Tab.~\ref{tab:overexposure_sat_ratio} shows that \textit{i-PhysGaussian} maintains \textbf{low saturation} on both real-captured scenes and closely matches PhysGaussian (mean Sat.\%: $0.15$ on Alocasia and $0.05$ on Hat), indicating that introducing an implicit MPM solver does \emph{not} cause systematic over-exposure degradation.
In contrast, Physics3D exhibits extremely high saturation means (Alocasia: $98.14$, Hat: $92.93$) with near-zero variability, consistent with saturation-dominated degenerate outputs.
Overall, our pipeline preserves exposure stability on visually complex real scenes and remains comparable to strong explicit-MPM baselines.

\section{Conclusion}
\label{sec:col}
In summary, \textit{i-PhysGaussian} tightly couples 3D Gaussian Splatting with a fully implicit MPM integrator by enforcing within-step momentum balance and solving the end-of-step state via Newton–GMRES, substantially mitigating the step-size sensitivity and failure modes of explicit methods under stiff, highly nonlinear, and contact-rich dynamics; experiments demonstrate consistent improvements in large-step stability, trajectory-level robustness to time-step enlargement, and reference-free physical-consistency diagnostics, while preserving competitive rendering fidelity and exposure stability on real, texture-rich scenes. These robustness gains, however, come at the cost of higher per-step computation due to the nested implicit solver, particularly in challenging regimes, motivating future work on streamlining the solver through stronger preconditioning and warm starts, adaptive inexact solves with early-exit criteria, and reuse of intermediate quantities, to bridge the efficiency gap while retaining large-step stability for scaling to larger scenes and longer-horizon simulations.

\bibliographystyle{unsrt}
\bibliography{references}

\appendix
\clearpage

\begin{center}
 \rule{6.50in}{1.2pt}\\
 {\Large\bf Appendix for ``i-PhysGaussian:\\ \vspace{0.06in} Implicit Physical Simulation for 3D Gaussian Splatting''}
 \rule{6.50in}{1.2pt}
\end{center}

In the appendix, we first provide extended background on scene reconstruction and Material Point Methods (Section~\ref{sec:related_work}). 
We then summarize the hardware setup and the per-scene time-step schedules used for all methods (Section~\ref{App:repro}), including the impulse rescaling rule for the one-substep excitation in \textsc{Ficus}. 
Next, we detail the definitions of our stability and time-step robustness metrics and report additional results (Sections~\ref{App:stability_further}--\ref{App:AUC}). 
We further provide reference-free physical-consistency diagnostics (mass drift and impulse/torque irregularity) with additional qualitative overlays and per-scene discussions (Sections~\ref{App:MassDrift}--\ref{App:PhysPlausPerScene}). 
Finally, we include additional rendering-quality diagnostics and ablations on key components of the Newton--GMRES solver (Sections~\ref{App:Exposure}--\ref{tab:fixed_forcing_ablation}).

\section{Related Work}
\label{sec:related_work}

\subsection{Scene Reconstruction}

Neural scene representations have emerged as the dominant paradigm for high-fidelity scene reconstruction and novel view synthesis. NeRF~\cite{Mildenhall2020NeRF} models a continuous 5D radiance field and optimizes a multilayer perceptron (MLP) with differentiable volume rendering, achieving impressive view consistency and fine geometric detail. However, NeRF-style representations are typically implicit—encoding the scene within network weights—which often results in computationally intensive training and rendering. While extensive follow-up research has improved efficiency via factorization, acceleration structures, or alternative parameterizations~\cite{Yu2021PlenOctrees, Reiser2021KiloNeRF, Garbin2021FastNeRF, Mueller2022InstantNGP, FridovichKeil2022Plenoxels, Sun2022DVGO}, the learned scene content generally remains implicit. This characteristic limits direct inspection, editing, and seamless integration into downstream modules.

To achieve an explicit and directly renderable representation, recent studies have explored point-based and primitive-based scene models. 3D Gaussian Splatting (3DGS)~\cite{Kerbl_2023_3DGS} represents a scene as a set of anisotropic Gaussians with low-order spherical harmonics and visibility-aware splatting, enabling real-time rendering alongside high-quality reconstructions. Crucially, 3DGS provides an explicit, editable representation that can be saved and manipulated without decoding a neural network. This makes it particularly well-suited as a backbone for physics-based dynamic simulation, where the reconstructed scene must interface directly with a mechanics solver. This explicitness also simplifies exporting geometry and state data to physics solvers and facilitates the re-rendering of simulated states, which is essential for reconstruction-driven dynamics.

\subsection{Material Point Methods}

The Material Point Method (MPM), introduced by Sulsky et al.~\cite{Sulsky1994MPM}, is a hybrid particle–grid framework for continuum mechanics: Lagrangian material points carry the state, while an Eulerian background grid facilitates the updates. This design avoids mesh entanglement during large deformations and preserves the computational efficiency of regular grids.

Transfers between particles and the grid can introduce interpolation errors. To mitigate these effects, Bardenhagen and Kober proposed the Generalized Interpolation Material Point (GIMP) method~\cite{Bardenhagen2004GIMP}, which assigns particles a finite domain and smooths transfer weights across cell boundaries. Subsequent developments, such as APIC~\cite{Jiang2015APIC} and MLS-MPM~\cite{Hu2018MLSPMM}, have further improved linear/angular momentum conservation and the stability of P2G/G2P transfers.

For time integration, while explicit schemes remain widely used due to their simplicity, semi-implicit~\cite{Daviet2016SemiImplicitGranularMPM, KularathnaSoga2017ImplicitIncompressibleMPM} and fully implicit variants~\cite{GuilkeyWeiss2003ImplicitMPM, SulskyKaul2004ImplicitDynamicsMPM,huang2023robust,huang2023flatmatch} have long been established in traditional simulation. These variants improve robustness for stiff materials and enable stable roll-outs with larger time steps. Despite these advantages, implicit MPM remains \emph{underexplored} in reconstruction-driven dynamic pipelines (\textit{e.g.}, those built on NeRF or 3DGS), where most methods still rely on explicit stepping for advancing dynamics.

More recently, differentiable MPM~\cite{Hu2019ChainQueen, Hu2020DiffTaichi} has leveraged automatic differentiation to provide end-to-end gradients for control, parameter identification, and learning-based simulation, further broadening the applicability of MPM in data-driven settings.

\newpage
\section{Reproducibility and Hardware.}
\label{App:repro}
We follow a unified evaluation protocol across methods and keep camera settings and per-scene configurations consistent whenever applicable.
All experiments are conducted on a single NVIDIA RTX 4090 GPU (24GB).

Tables~\ref{tab:ficus_config}--\ref{tab:bread_config} summarize the time-step schedules for the \textsc{Ficus}, \textsc{Pillow2Sofa}, and \textsc{Bread} scenes (and the impulse boundary condition for \textsc{Ficus}).
We sweep the time-step multiplier $k\in\{1,2,4,\ldots,20\}$ by scaling the substep size as
$\texttt{substep\_dt}(k)=k\cdot \texttt{substep\_dt}@1\times\Delta t$ while keeping \texttt{frame\_dt} and \texttt{frame\_num} fixed.
Since $\texttt{frame\_dt}/\texttt{substep\_dt}(k)$ is not always an integer, we follow the implementation in our code and set
\[
\texttt{step\_per\_frame}=\mathrm{round}\!\left(\frac{\texttt{frame\_dt}}{\texttt{substep\_dt}(k)}\right),
\]
and advance each frame using exactly \texttt{step\_per\_frame} substeps of size \texttt{substep\_dt}(k).
Notably, the \textsc{Ficus} scene includes a \texttt{particle\_impulse} boundary condition that acts for exactly one substep. 
To ensure a consistent total impulse across different $k$, we scale the force magnitude proportionally as $1/k$.

In addition, \texttt{Physics3D} and \texttt{DreamPhysics} adopt a different numerical convention for Young's modulus $E$ in their configuration files: $E_{\text{cfg}}=1.0$ corresponds to a physical modulus of $E_{\text{phys}}=10^{7}$.
To align material settings across methods, we apply the conversion
\[
E_{\text{cfg}}=\frac{E_{\text{phys}}}{10^{7}}
\quad\text{(equivalently, } E_{\text{phys}}=10^{7}E_{\text{cfg}}\text{)},
\]
and report all $E$ values in physical units elsewhere unless stated otherwise.
We therefore convert and re-scale $E$ according to this mapping to align the material settings across methods. For more implementation details and scene-specific parameters, please refer to the configuration files included in our release.

\begin{table}[t]
\caption{\textbf{Ficus} scene: time-step schedule and particle-impulse boundary condition.
We sweep $k\in\{1,2,4,6,8,10,12,14,16,18,20\}$ by setting \texttt{substep\_dt}=$k\times 10^{-4}$ while keeping \texttt{frame\_dt}=$4.0\mathrm{e}{-2}$ and \texttt{frame\_num}=125 fixed. The particle impulse uses \texttt{force}=\texttt{[}$f_x$,0,0\texttt{]}, \texttt{num\_dt}=1, and \texttt{start\_time}=0.0.}
\label{tab:ficus_config}
\centering
\small
\begin{sc}
\setlength{\tabcolsep}{0.5pt}
\begin{tabular}{c ccc ccc}
\toprule
& \multicolumn{3}{c}{\textbf{time}} & \multicolumn{3}{c}{\textbf{boundary\_conditions: particle\_impulse}} \\
\cmidrule(lr){2-4}\cmidrule(lr){5-7}
\textbf{time-size} & \texttt{substep\_dt} & \texttt{frame\_dt} & \texttt{frame\_num} & $f_x$ & \texttt{num\_dt} & \texttt{start\_time} \\
\midrule
$1\times \Delta t$  & $1.0\mathrm{e}{-4}$ & $4.0\mathrm{e}{-2}$ & 125 & $-1.8\mathrm{e}{-1}$     & 1 & 0.0 \\
$2\times \Delta t$  & $2.0\mathrm{e}{-4}$ & $4.0\mathrm{e}{-2}$ & 125 & $-9.0\mathrm{e}{-2}$     & 1 & 0.0 \\
$4\times \Delta t$  & $4.0\mathrm{e}{-4}$ & $4.0\mathrm{e}{-2}$ & 125 & $-4.5\mathrm{e}{-2}$     & 1 & 0.0 \\
$6\times \Delta t$  & $6.0\mathrm{e}{-4}$ & $4.0\mathrm{e}{-2}$ & 125 & $-3.0\mathrm{e}{-2}$     & 1 & 0.0 \\
$8\times \Delta t$  & $8.0\mathrm{e}{-4}$ & $4.0\mathrm{e}{-2}$ & 125 & $-2.25\mathrm{e}{-2}$    & 1 & 0.0 \\
$10\times \Delta t$ & $1.0\mathrm{e}{-3}$ & $4.0\mathrm{e}{-2}$ & 125 & $-1.8\mathrm{e}{-2}$     & 1 & 0.0 \\
$12\times \Delta t$ & $1.2\mathrm{e}{-3}$ & $4.0\mathrm{e}{-2}$ & 125 & $-1.5\mathrm{e}{-2}$     & 1 & 0.0 \\
$14\times \Delta t$ & $1.4\mathrm{e}{-3}$ & $4.0\mathrm{e}{-2}$ & 125 & $-1.28571\mathrm{e}{-2}$ & 1 & 0.0 \\
$16\times \Delta t$ & $1.6\mathrm{e}{-3}$ & $4.0\mathrm{e}{-2}$ & 125 & $-1.125\mathrm{e}{-2}$   & 1 & 0.0 \\
$18\times \Delta t$ & $1.8\mathrm{e}{-3}$ & $4.0\mathrm{e}{-2}$ & 125 & $-1.0\mathrm{e}{-2}$     & 1 & 0.0 \\
$20\times \Delta t$ & $2.0\mathrm{e}{-3}$ & $4.0\mathrm{e}{-2}$ & 125 & $-9.0\mathrm{e}{-3}$     & 1 & 0.0 \\
\bottomrule
\end{tabular}
\end{sc}
\end{table}

\begin{table}[t]
\caption{\textbf{Pillow2Sofa} scene: time-step schedule.
We sweep $k\in\{1,2,4,6,8,10,12,14,16,18,20\}$ by setting \texttt{substep\_dt}=$k\times 10^{-4}$ while keeping \texttt{frame\_dt}=$2.0\mathrm{e}{-2}$ and \texttt{frame\_num}=300 fixed.}

\label{tab:pillow2sofa_config}
\centering
\small
\begin{sc}
\setlength{\tabcolsep}{0.5pt}
\begin{tabular}{cccc}
\toprule
& \multicolumn{3}{c}{\textbf{time}} \\
\cmidrule(lr){2-4}
\textbf{time-size} & \texttt{substep\_dt} & \texttt{frame\_dt} & \texttt{frame\_num} \\
\midrule
$1\times \Delta t$  & $1.0\mathrm{e}{-4}$ & $2.0\mathrm{e}{-2}$ & 300 \\
$2\times \Delta t$  & $2.0\mathrm{e}{-4}$ & $2.0\mathrm{e}{-2}$ & 300 \\
$4\times \Delta t$  & $4.0\mathrm{e}{-4}$ & $2.0\mathrm{e}{-2}$ & 300 \\
$6\times \Delta t$  & $6.0\mathrm{e}{-4}$ & $2.0\mathrm{e}{-2}$ & 300 \\
$8\times \Delta t$  & $8.0\mathrm{e}{-4}$ & $2.0\mathrm{e}{-2}$ & 300 \\
$10\times \Delta t$ & $1.0\mathrm{e}{-3}$ & $2.0\mathrm{e}{-2}$ & 300 \\
$12\times \Delta t$ & $1.2\mathrm{e}{-3}$ & $2.0\mathrm{e}{-2}$ & 300 \\
$14\times \Delta t$ & $1.4\mathrm{e}{-3}$ & $2.0\mathrm{e}{-2}$ & 300 \\
$16\times \Delta t$ & $1.6\mathrm{e}{-3}$ & $2.0\mathrm{e}{-2}$ & 300 \\
$18\times \Delta t$ & $1.8\mathrm{e}{-3}$ & $2.0\mathrm{e}{-2}$ & 300 \\
$20\times \Delta t$ & $2.0\mathrm{e}{-3}$ & $2.0\mathrm{e}{-2}$ & 300 \\
\bottomrule
\end{tabular}
\end{sc}
\end{table}

\begin{table}[t]
\caption{\textbf{Bread} scene: time-step schedule.
We sweep $k\in\{1,2,4,6,8,10,12,14,16,18,20\}$ by setting \texttt{substep\_dt}=$k\times 10^{-4}$ while keeping \texttt{frame\_dt}=$1.0\mathrm{e}{-2}$ and \texttt{frame\_num}=300 fixed.}

\label{tab:bread_config}
\centering
\small
\begin{sc}
\setlength{\tabcolsep}{0.5pt}
\begin{tabular}{cccc}
\toprule
& \multicolumn{3}{c}{\textbf{time}} \\
\cmidrule(lr){2-4}
\textbf{time-size} & \texttt{substep\_dt} & \texttt{frame\_dt} & \texttt{frame\_num} \\
\midrule
$1\times \Delta t$  & $1.0\mathrm{e}{-4}$ & $1.0\mathrm{e}{-2}$ & 300 \\
$2\times \Delta t$  & $2.0\mathrm{e}{-4}$ & $1.0\mathrm{e}{-2}$ & 300 \\
$4\times \Delta t$  & $4.0\mathrm{e}{-4}$ & $1.0\mathrm{e}{-2}$ & 300 \\
$6\times \Delta t$  & $6.0\mathrm{e}{-4}$ & $1.0\mathrm{e}{-2}$ & 300 \\
$8\times \Delta t$  & $8.0\mathrm{e}{-4}$ & $1.0\mathrm{e}{-2}$ & 300 \\
$10\times \Delta t$ & $1.0\mathrm{e}{-3}$ & $1.0\mathrm{e}{-2}$ & 300 \\
$12\times \Delta t$ & $1.2\mathrm{e}{-3}$ & $1.0\mathrm{e}{-2}$ & 300 \\
$14\times \Delta t$ & $1.4\mathrm{e}{-3}$ & $1.0\mathrm{e}{-2}$ & 300 \\
$16\times \Delta t$ & $1.6\mathrm{e}{-3}$ & $1.0\mathrm{e}{-2}$ & 300 \\
$18\times \Delta t$ & $1.8\mathrm{e}{-3}$ & $1.0\mathrm{e}{-2}$ & 300 \\
$20\times \Delta t$ & $2.0\mathrm{e}{-3}$ & $1.0\mathrm{e}{-2}$ & 300 \\
\bottomrule
\end{tabular}
\end{sc}
\end{table}

\begin{table}[t]
\caption{\textbf{Material scope and constitutive models.}}
\label{tab:material_constitutive_mapping}
\centering
\small
\begin{sc}
\setlength{\tabcolsep}{6pt}
\begin{tabular}{lll}
\toprule
Material & Type & Model \\
\midrule
\texttt{jelly}      & Elastic      & Neo-Hookean \\
\texttt{metal}      & Plastic      & J2 (von Mises) \\
\texttt{foam}       & Damage       & J2 + softening \\
\texttt{snow/sand}  & Frictional   & Drucker--Prager \\
\texttt{plasticine} & Viscoplastic & Rate-dependent J2 \\
\bottomrule
\end{tabular}
\end{sc}
\end{table}

Table~\ref{tab:material_constitutive_mapping} summarizes the material presets used in our experiments and their underlying constitutive formulations, spanning hyperelasticity (Neo-Hookean), elastoplasticity (Hencky elasticity with J2/von~Mises yield), frictional granular behavior (Drucker--Prager-type plasticity), and rate-dependent viscoplasticity.

\newpage
\section{Stability Metrics and Additional Results}
\label{App:stability_further}

\subsection{Boundary-Hit Mass Fraction (BMF) definition}
\label{App:BMF}
At frame $t$, BMF is defined as the fraction of total particle mass belonging to collapsed particles,
\[
\mathrm{BMF}_t \;=\; \frac{\sum_{i\in\mathcal{C}_t} m_i}{\sum_i m_i},
\]
where $\mathcal{C}_t$ denotes the set of particles that were clamped during the step that produces frame $t$.\footnote{Concretely, a particle is included in $\mathcal{C}_t$ if any coordinate is clipped to $\epsilon$ or $\texttt{grid\_lim}-\epsilon$ by the clamp rule.}

\subsection{Additional Heatmap Details}

% -------------------- Figure 2: Stability heatmaps --------------------
\begin{figure}[t]
    \centering
    \includegraphics[width=\linewidth]{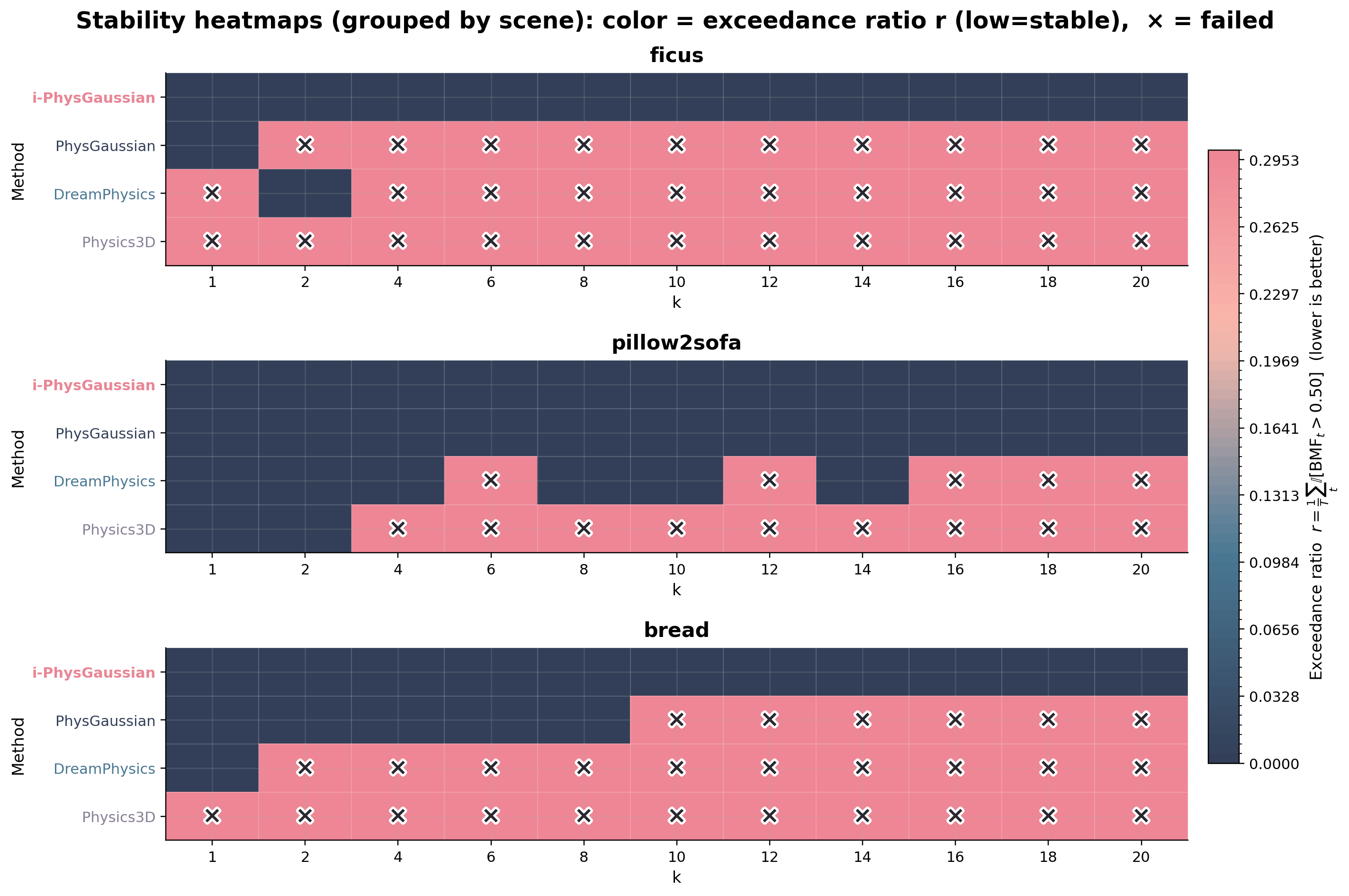}
    \caption{\textbf{Stability heatmaps} across time-step multipliers. Color indicates exceedance ratio
    $r=\frac{1}{T}\sum_{t=1}^{T}\mathbb{I}[\mathrm{BMF}_t>0.5]$ (lower is better); $\times$ marks runs that fail the gate ($r>0.5$).}
    \label{fig:bmf_stability_heatmaps}
\end{figure}

Fig.~\ref{fig:bmf_stability_heatmaps} further offers a fine-grained characterization of stability via a per-setting heatmap. 
For each (scene, method, $k$), the cell color encodes the exceedance ratio
\[
r \;=\; \frac{1}{T}\sum_{t=1}^{T}\mathbb{I}\!\left[\mathrm{BMF}_t>0.5\right],
\]
\textit{i.e.} the fraction of frames in which BMF exceeds the threshold (lower is more stable), and we overlay ``$\times$'' to mark settings that are deemed \emph{failed} under the unified BMF-gate rule. 
Consistent with the bar-chart summary, \textit{i-PhysGaussian} maintains low $r$ and never triggers failure across all scenes and multipliers. 
PhysGaussian enters the high-$r$ regime earlier on Bread as $k$ increases, whereas DreamPhysics and Physics3D show earlier and broader regions of high $r$ on Pillow2Sofa and Bread, which explains their higher failure rates and smaller $k_{\max}$.

\newpage
\section{Time-step Robustness Metrics and Additional Results}

\subsection{Center-of-mass drift (COMD) definition}
\label{App:COMD}

Let $\mathbf{x}^{(k)}_{t,i}\in\mathbb{R}^3$ be the position of particle $i$ at frame $t$ under the $k\times\Delta t$ setting, with mass $m_i$.
Let $\mathbf{x}^{(1)}_{t,i}$ denote the $1\times\Delta t$ reference.
We first compute the mass-weighted center of mass at each frame:
\begin{equation}
\mathbf{c}^{(k)}_t=\frac{\sum_i m_i\mathbf{x}^{(k)}_{t,i}}{\sum_i m_i}.
\end{equation}
We then define COMD as the time-averaged normalized displacement to the reference:
\begin{equation}
\mathrm{COMD}^{(k)} \;=\; \frac{1}{T}\sum_{t=1}^{T}\frac{1}{\texttt{grid\_lim}}
\left\|\mathbf{c}^{(k)}_t-\mathbf{c}^{(1)}_t\right\|_2.
\end{equation}

\subsection{Mass-weighted RMS Drift (mwRMSD) with clamping penalty}
\label{App:mwRMSD}

To ensure hard clamping does not artificially suppress divergence (by projecting out-of-domain motion back to the boundary), we impose a conservative penalty on particles that are clamped in either trajectory.
Let $\mathcal{C}^{(k)}_t$ and $\mathcal{C}^{(1)}_t$ denote the sets of particles that are clamped during the simulation step producing frame $t$ under $k\times$ and $1\times$, respectively, and define
\begin{equation}
\mathcal{C}_t=\mathcal{C}^{(k)}_t\cup\mathcal{C}^{(1)}_t.
\end{equation}
We compute the per-particle deviation:
\begin{equation}
d^{(k)}_{t,i}=\left\|\mathbf{x}^{(k)}_{t,i}-\mathbf{x}^{(1)}_{t,i}\right\|_2,
\end{equation}
and define a penalized deviation with an upper bound $D_{\max}=\texttt{grid\_lim}$:
\begin{equation}
\hat d^{(k)}_{t,i}=
\begin{cases}
D_{\max}, & i\in \mathcal{C}_t,\\[2pt]
\min\!\left(d^{(k)}_{t,i},\,D_{\max}\right), & \text{otherwise.}
\end{cases}
\end{equation}
Finally, mwRMSD is defined as the time-averaged normalized mass-weighted RMS deviation:
\begin{equation}
\mathrm{mwRMSD}^{(k)} \;=\; \frac{1}{T}\sum_{t=1}^{T}\frac{1}{\texttt{grid\_lim}}
\sqrt{\frac{\sum_i m_i\left(\hat d^{(k)}_{t,i}\right)^2}{\sum_i m_i}}.
\end{equation}

\subsection{Additional AUC details}
\label{App:AUC}

To summarize robustness across multipliers with a single number, we report the AUC, which is computed over all tested multipliers $k\in{1,2,4,\dots,20}$, and should be interpreted together with the BMF gate since clamp-dominated regimes can distort drift magnitudes.
Lower AUC indicates smaller overall drift under time-step enlargement.

\begin{table}[t]
\caption{Normalized AUC of drift curves (lower is better), computed over $k\in\{1,2,4,\dots,20\}$ on BMF-valid runs.}
\label{tab:dt_robustness_auc}
\centering
\small
\setlength{\tabcolsep}{8pt}
\begin{sc}
\begin{tabular}{lcc}
\toprule
Method & COMD AUC$\downarrow$ & mwRMSD AUC$\downarrow$ \\
\midrule
\multicolumn{3}{c}{\textbf{Ficus}} \\
\textit{i-PhysGaussian} & 0.0184 & 0.0279 \\
PhysGaussian            & 0.5975 & 0.8509 \\
DreamPhysics            & 0.0556 & 0.8539 \\
Physics3D               & 0.0000 & 0.8539 \\
\midrule
\multicolumn{3}{c}{\textbf{Pillow2Sofa}} \\
\textit{i-PhysGaussian} & 0.0023 & 0.0382 \\
PhysGaussian            & 0.0139 & 0.0406 \\
DreamPhysics            & 0.4484 & 0.4756 \\
Physics3D               & 0.8169 & 0.8867 \\
\midrule
\multicolumn{3}{c}{\textbf{Bread}} \\
\textit{i-PhysGaussian} & 0.0138 & 0.2126 \\
PhysGaussian            & 0.3394 & 0.5202 \\
DreamPhysics            & 0.8359 & 0.9737 \\
Physics3D               & 0.0000 & 0.9737 \\
\bottomrule
\end{tabular}
\end{sc}
\end{table}

Overall, \textit{i-PhysGaussian} achieves the lowest (or near-lowest) AUC across scenes and metrics, consistent with the drift curves in Fig.~\ref{fig:dt_robustness_drift}.
PhysGaussian is close to \textit{i-PhysGaussian} on Pillow2Sofa but degrades on Bread, whereas DreamPhysics and Physics3D incur substantially larger AUC on the harder scenes.
We interpret COMD and mwRMSD jointly: COMD captures global motion bias while mwRMSD reflects particle-level deviation and deformation details.

\newpage
\section{Physical Simulation Fidelity Metrics and Additional Results}

\subsection{Mass drift definition}
\label{App:MassDrift}
Let $\mathbf{x}_{t,i}\in\mathbb{R}^3$ denote the position of particle $i$ at frame $t$ with mass $m_i$ under the $1\times\Delta t$ setting, for $t=1,\dots,T$. We define the total mass
\begin{equation}
M_t=\sum_i m_i,
\end{equation}
and report the per-frame mass drift relative to the first frame:
\begin{equation}
\mathrm{MassDrift}_t
\;=\;
\frac{\left|M_t-M_1\right|}{M_1+\varepsilon},
\quad t=1,\dots,T,
\end{equation}
where $\varepsilon$ is a small constant for numerical stability.

\subsection{Momentum and irregularity}
\label{App:PhysPlausMom}
We estimate particle velocities from trajectories via finite differences:
\begin{equation}
\mathbf{v}_{t,i}=
\begin{cases}
\frac{\mathbf{x}_{t+1,i}-\mathbf{x}_{t,i}}{\Delta t}, & t=1,\\[4pt]
\frac{\mathbf{x}_{t+1,i}-\mathbf{x}_{t-1,i}}{2\Delta t}, & 1<t<T,\\[4pt]
\frac{\mathbf{x}_{t,i}-\mathbf{x}_{t-1,i}}{\Delta t}, & t=T,
\end{cases}
\end{equation}
and compute the total linear momentum
\begin{equation}
\mathbf{P}_t=\sum_i m_i\,\mathbf{v}_{t,i}.
\end{equation}
We also compute the instantaneous center of mass
\begin{equation}
\mathbf{c}_t=\frac{\sum_i m_i\mathbf{x}_{t,i}}{\sum_i m_i},
\end{equation}
and define the total angular momentum about $\mathbf{c}_t$:
\begin{equation}
\mathbf{L}_t=\sum_i \left(\mathbf{x}_{t,i}-\mathbf{c}_t\right)\times\left(m_i\,\mathbf{v}_{t,i}\right).
\end{equation}

We measure the \emph{second-order temporal difference} of the total momenta:
\begin{equation}
\Delta^2\mathbf{P}_t
=
\mathbf{P}_t - 2\mathbf{P}_{t-1} + \mathbf{P}_{t-2},
\quad t=3,\dots,T,
\end{equation}
\begin{equation}
\Delta^2\mathbf{L}_t
=
\mathbf{L}_t - 2\mathbf{L}_{t-1} + \mathbf{L}_{t-2},
\quad t=3,\dots,T.
\end{equation}
We normalize them using fixed scales to avoid division instability in near-static phases:
\begin{equation}
\mathrm{ImpulseIrr}_t
=
\frac{\left\|\Delta^2\mathbf{P}_t\right\|_2}{P_{\mathrm{scale}}+\varepsilon},
\quad
\mathrm{TorqueIrr}_t
=
\frac{\left\|\Delta^2\mathbf{L}_t\right\|_2}{L_{\mathrm{scale}}+\varepsilon}.
\end{equation}
Here we use the domain-scale, time-step-aware normalization
\begin{equation}
P_{\mathrm{scale}} = M_{\mathrm{total}}\,\frac{\texttt{grid\_lim}}{\Delta t},
\quad
L_{\mathrm{scale}} = M_{\mathrm{total}}\,\frac{\texttt{grid\_lim}^2}{\Delta t},
\end{equation}
where $M_{\mathrm{total}}=\sum_i m_i$ is the total mass.

Lower values indicate smoother (less jittery) temporal evolution of net impulse/torque, while sharp spikes suggest bursty non-physical impulses or numerical artifacts (e.g., due to hard clamping projection or unstable contact handling).

\subsection{Mass conservation sanity check}
\label{App:MassSanity}

\begin{figure*}[t]
    \centering
    \begin{subfigure}[t]{0.32\textwidth}
        \centering
        \includegraphics[width=\linewidth]{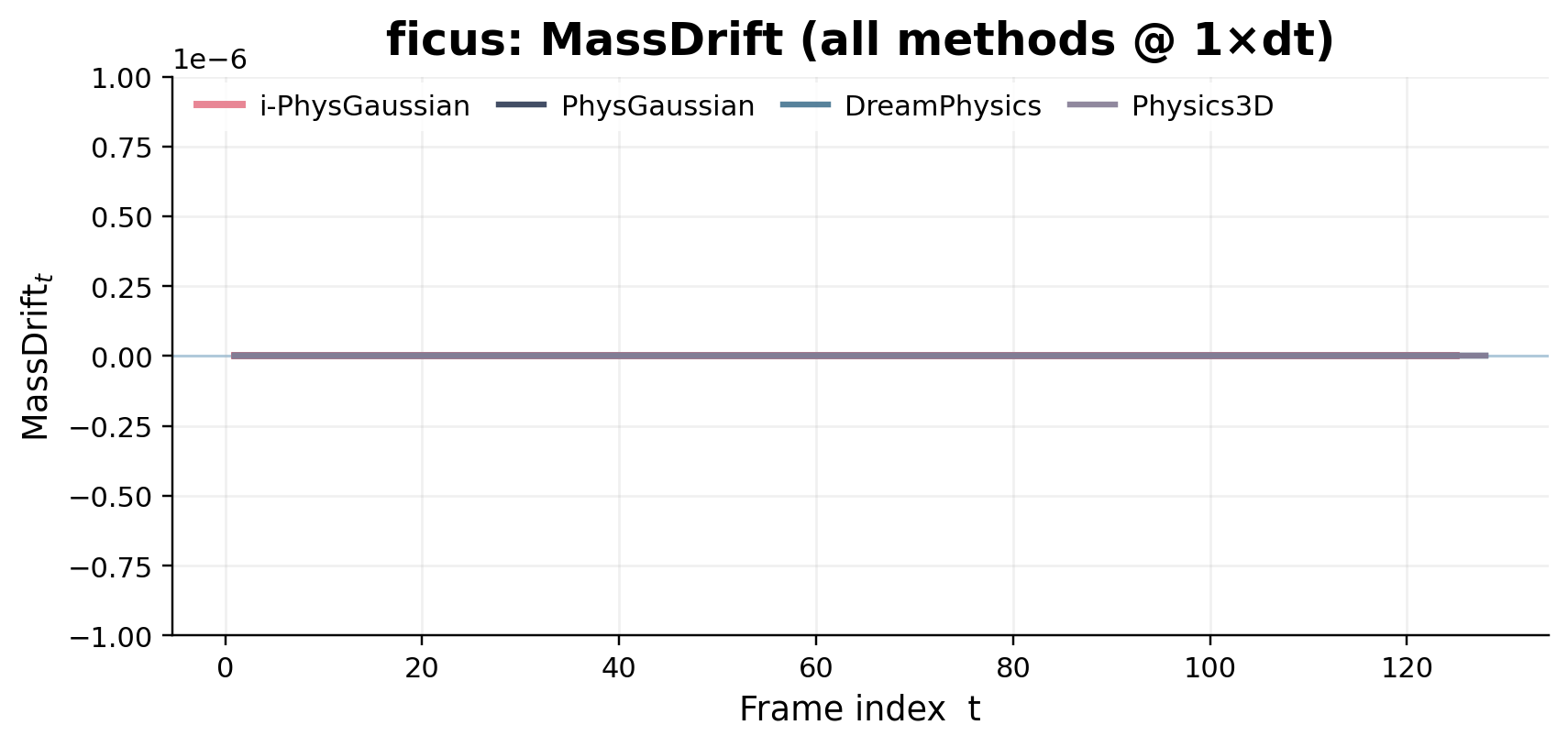}
        \caption{\textbf{Ficus}}
        \label{fig:massdrift_ficus}
    \end{subfigure}\hfill
    \begin{subfigure}[t]{0.32\textwidth}
        \centering
        \includegraphics[width=\linewidth]{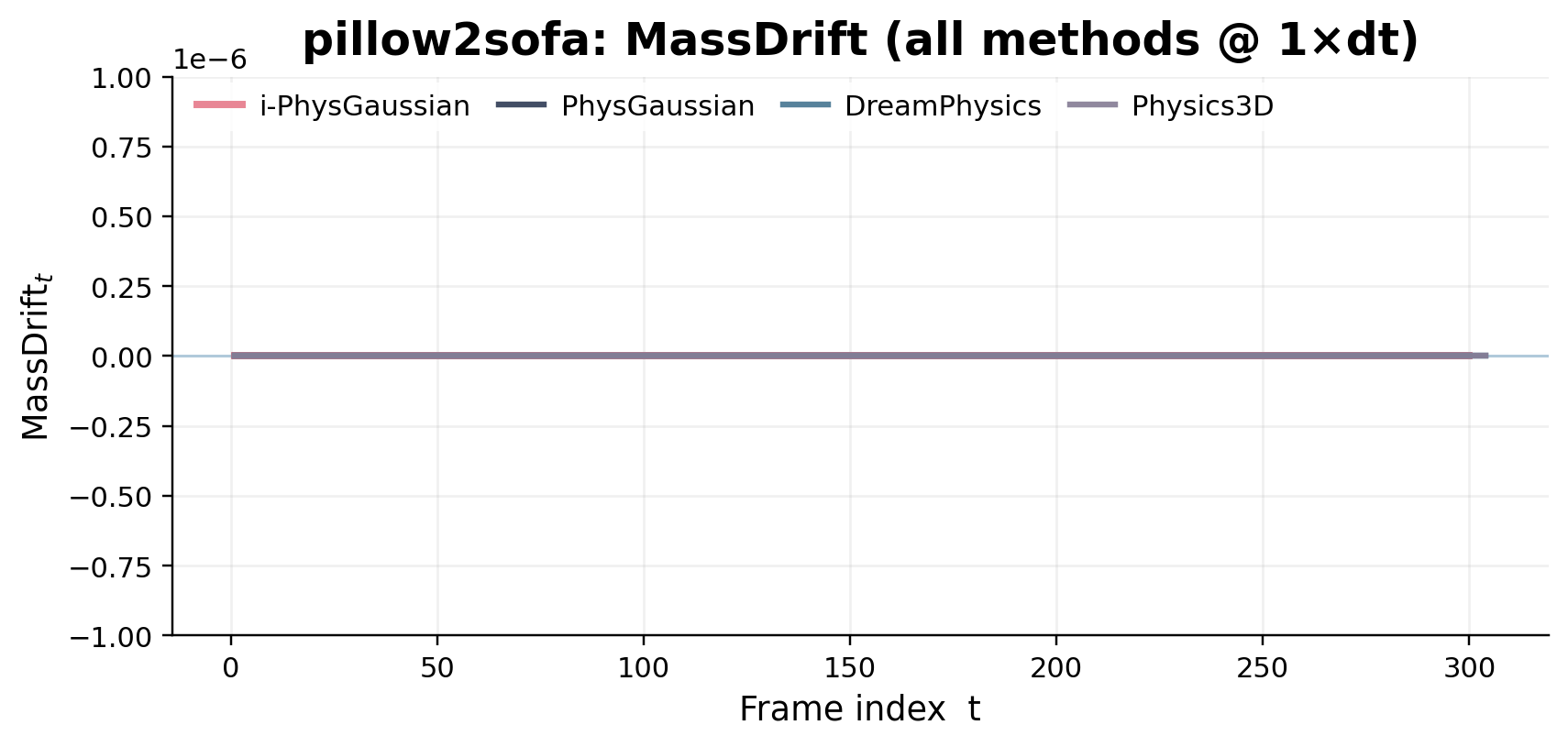}
        \caption{\textbf{Pillow2Sofa}}
        \label{fig:massdrift_pillow2sofa}
    \end{subfigure}\hfill
    \begin{subfigure}[t]{0.32\textwidth}
        \centering
        \includegraphics[width=\linewidth]{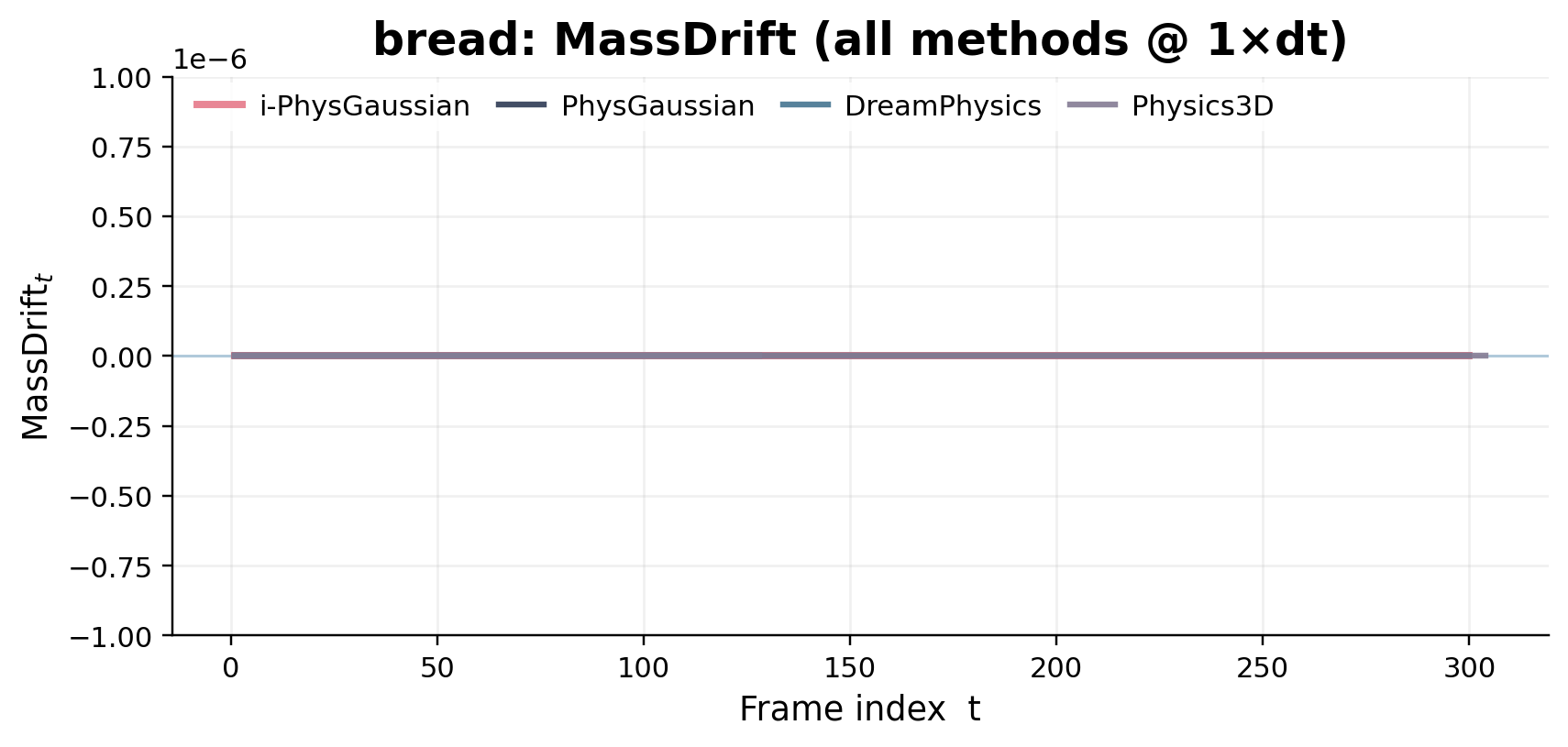}
        \caption{\textbf{Bread}}
        \label{fig:massdrift_bread}
    \end{subfigure}

    \caption{\textbf{Reference-free mass conservation at $1\times\Delta t$ (per-frame).}
    We plot MassDrift$_t=\frac{|M_t-M_1|}{M_1+\varepsilon}$ over frames for each scene (lower is better).}
    \label{fig:mass_consistency_1xdt}
\end{figure*}

\subsection{Additional per-scene discussion}
\label{App:PhysPlausPerScene}

% ===================== Key-trajectory visualization (3 scenes x 4 methods) =====================
% ===== single-column version (will be small) =====
\begin{figure*}[t]
    \centering
    \setlength{\tabcolsep}{1.5pt}
    \renewcommand{\arraystretch}{0.85}

    \newcommand{\trajimg}[2]{%
        \includegraphics[width=0.26\textwidth]{#1}%
        \phantomsubcaption\label{#2}%
    }

    \begin{tabular}{c c c c}
        & \textbf{Ficus} & \textbf{Pillow2Sofa} & \textbf{Bread} \\
        \midrule
        \rotatebox{90}{\hspace{1mm}\textbf{i-PhysGaussian}\hspace{1mm}} &
        \trajimg{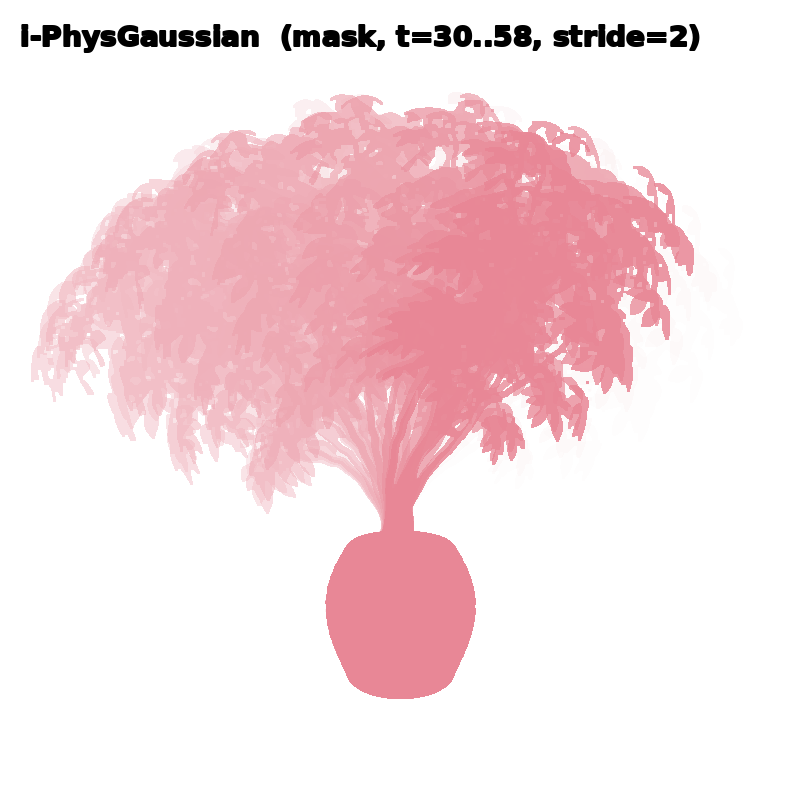}{fig:traj_ficus_ipg} &
        \trajimg{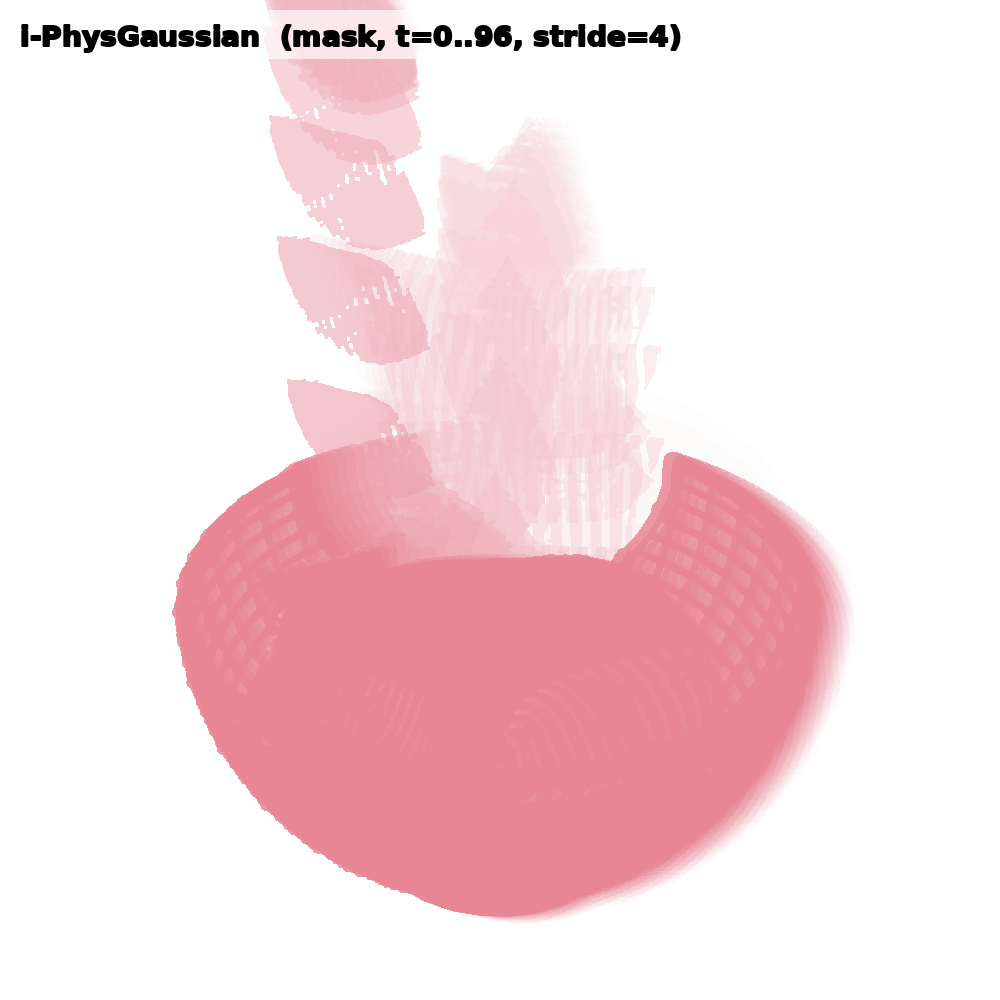}{fig:traj_pillow_ipg} &
        \trajimg{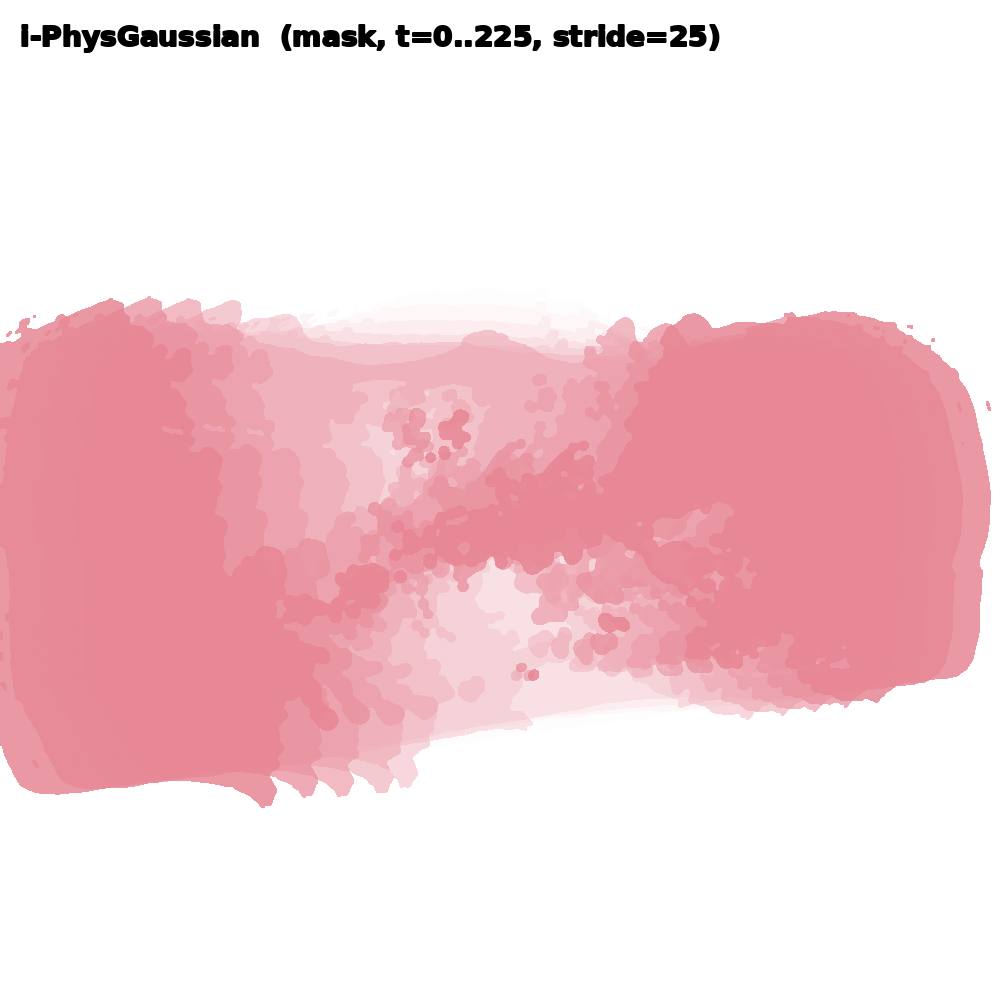}{fig:traj_bread_ipg} \\

        \rotatebox{90}{\hspace{1mm}\textbf{PhysGaussian}\hspace{1mm}} &
        \trajimg{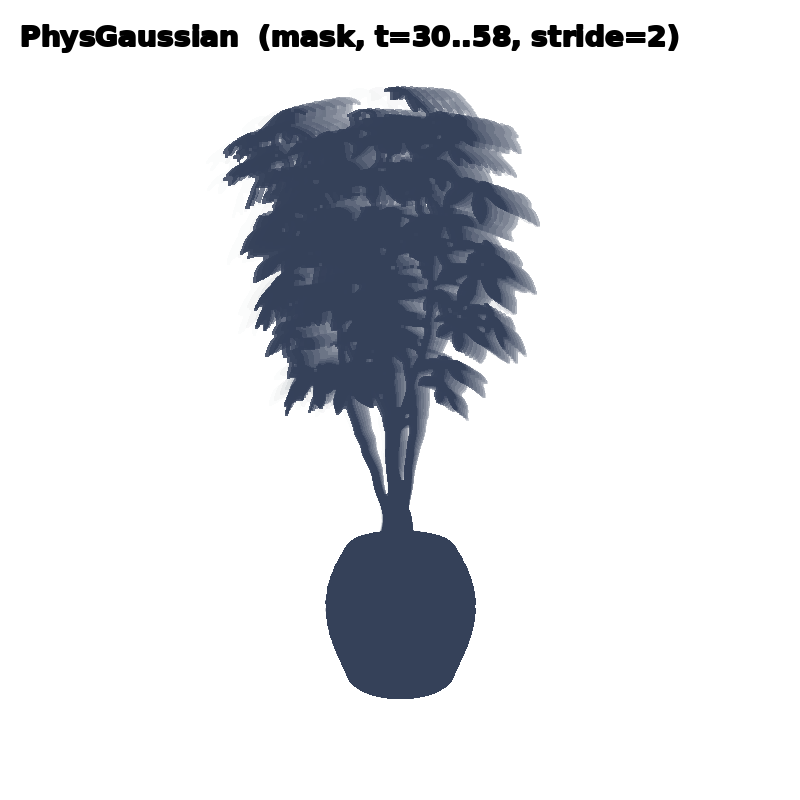}{fig:traj_ficus_pg} &
        \trajimg{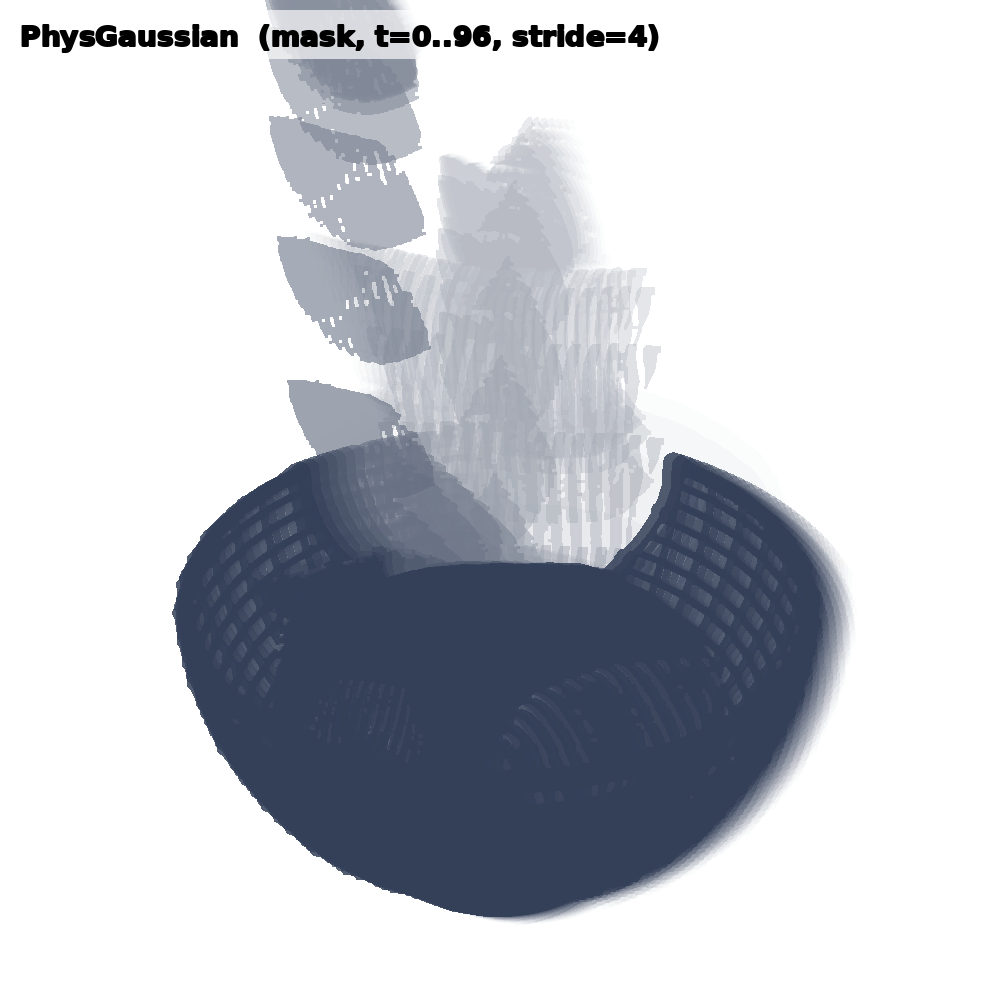}{fig:traj_pillow_pg} &
        \trajimg{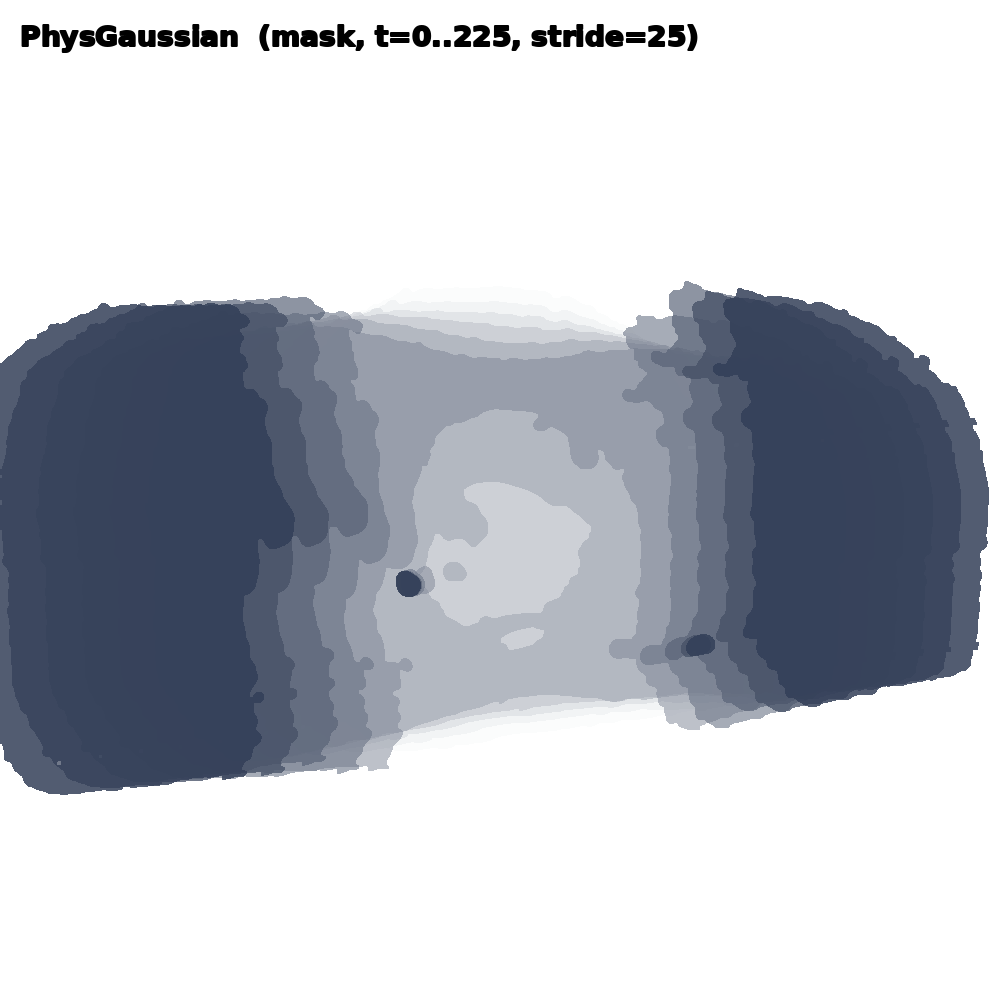}{fig:traj_bread_pg} \\

        \rotatebox{90}{\hspace{1mm}\textbf{DreamPhysics}\hspace{1mm}} &
        \trajimg{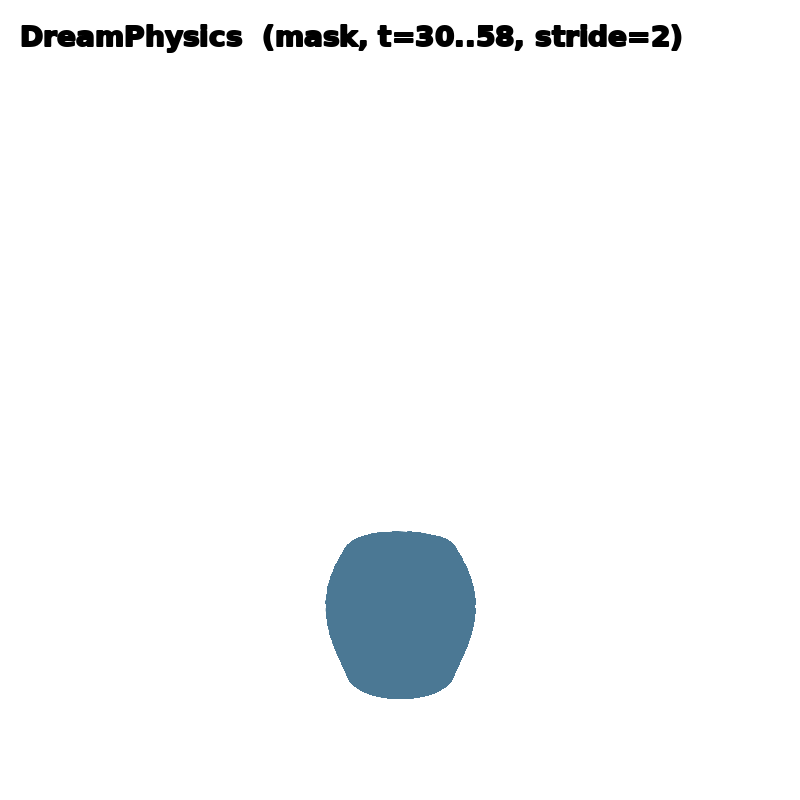}{fig:traj_ficus_dp} &
        \trajimg{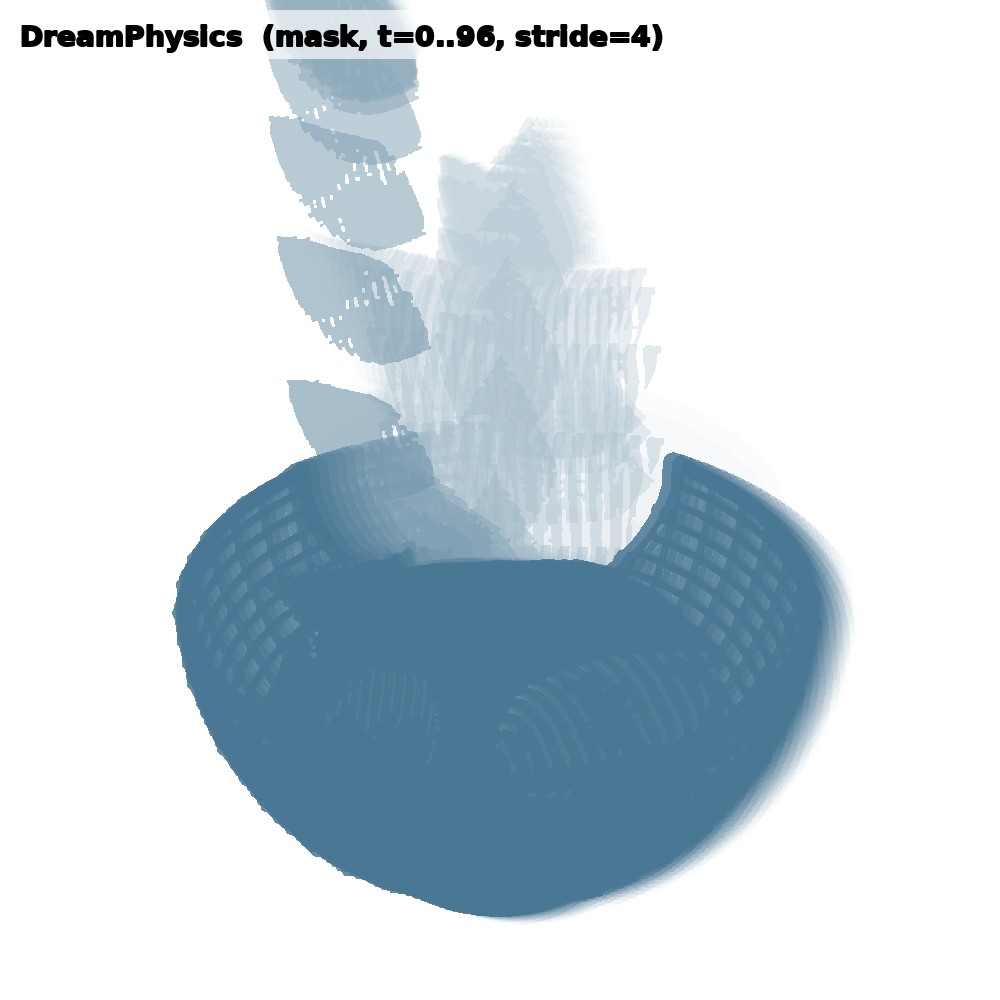}{fig:traj_pillow_dp} &
        \trajimg{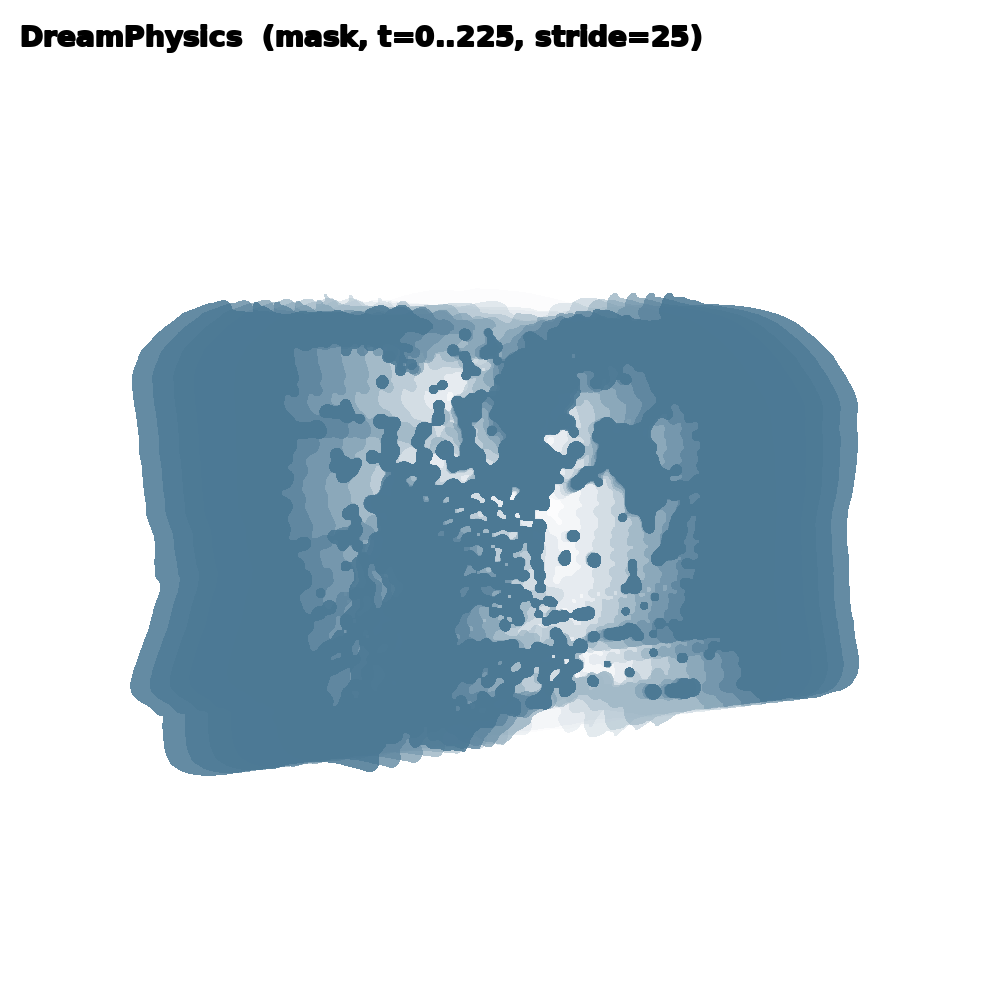}{fig:traj_bread_dp} \\

        \rotatebox{90}{\hspace{1mm}\textbf{Physics3D}\hspace{1mm}} &
        \trajimg{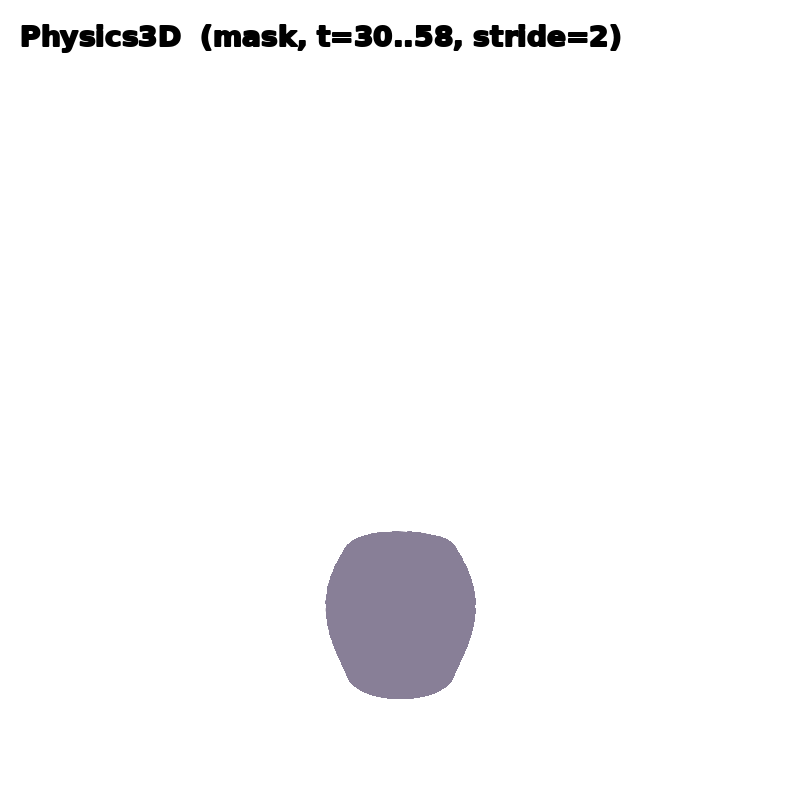}{fig:traj_ficus_p3d} &
        \trajimg{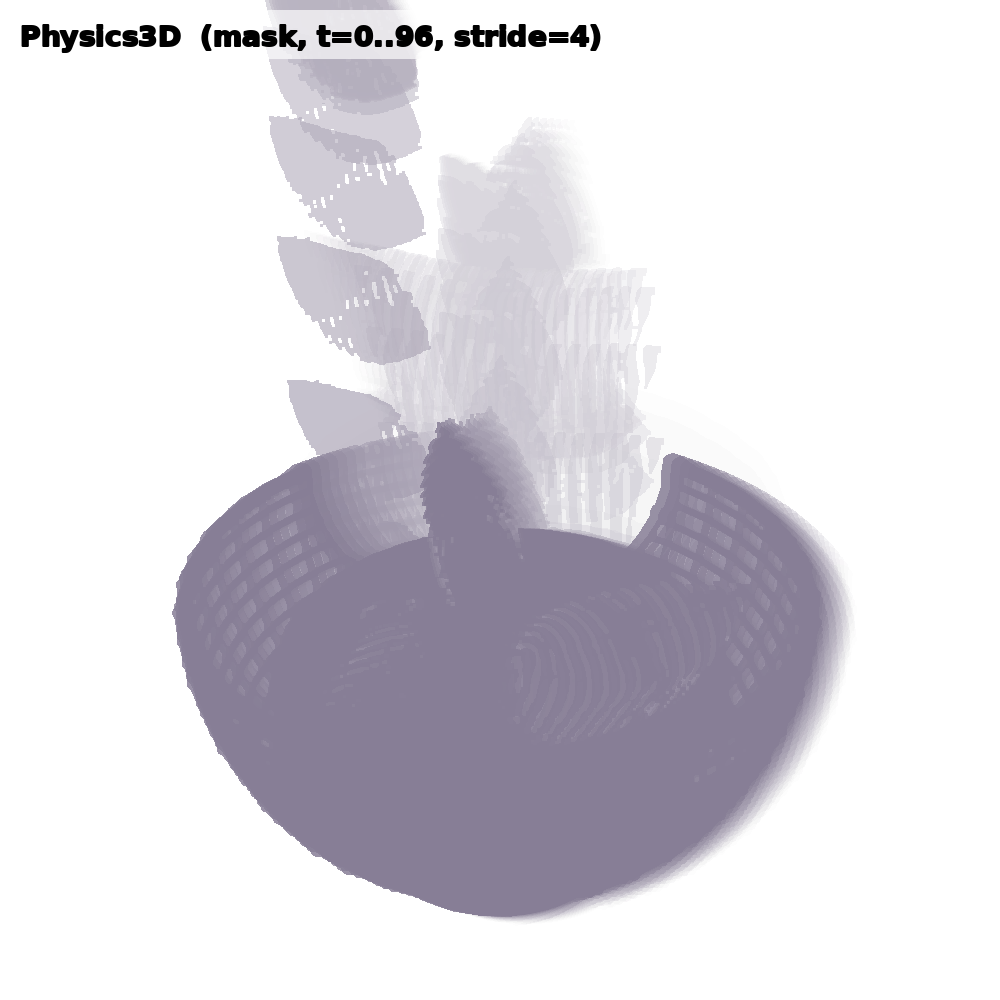}{fig:traj_pillow_p3d} &
        \trajimg{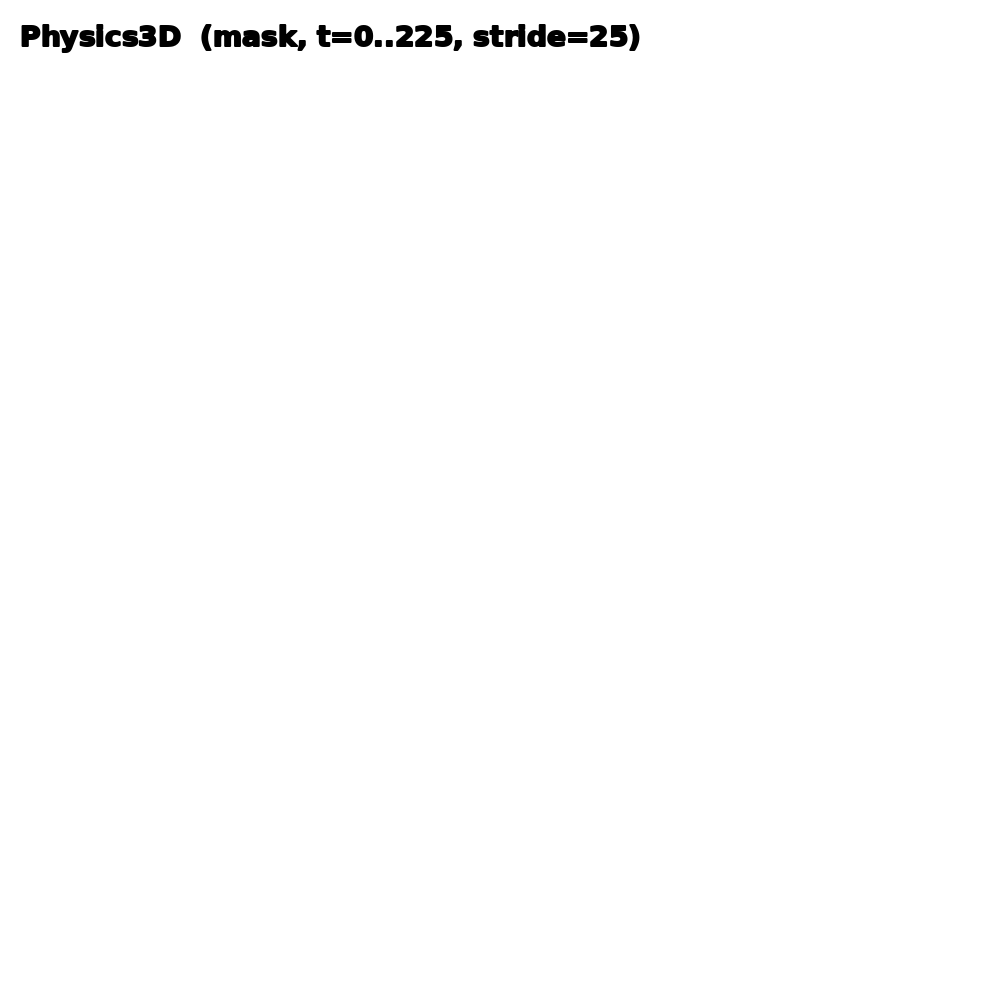}{fig:traj_bread_p3d} \\
    \end{tabular}

    \caption{\textbf{Trajectory overlays at a key simulation moment.}
    Near-flat irregularity curves can correspond to \emph{kinematic degeneration} (suppressed motion), as visualized here.}
    \label{fig:key_traj_overlays}
\end{figure*}

\paragraph{Ficus.}
Impulse/torque irregularities of \textit{i-PhysGaussian} and PhysGaussian show periodic, wave-like patterns consistent with the oscillatory swaying motion driven by external excitation.
Some baselines yield nearly flat curves close to zero; together with the trajectory overlays in Appendix Fig.~\ref{fig:key_traj_overlays} suggest kinematic degeneration toward an almost static trajectory.

\paragraph{Pillow2Sofa.}
All methods show well-aligned peaks in impulse/torque irregularity, corresponding to key contact/collision moments between the pillow and the sofa.
The trajectory overlays indicate broadly plausible collision dynamics across methods in this scene.

\paragraph{Bread.}
Overall, \textit{i-PhysGaussian} exhibits smoother curves with fewer bursty spikes.
PhysGaussian shows more abrupt spikes, consistent with rapid structural/contact-state changes.
Some baselines again exhibit near-zero irregularities, consistent with degeneration-to-static behavior as indicated by the trajectory overlays.

\newpage
\section{Dynamic Rendering Quality Metrics and Additional Results}
\label{App:Exposure}

\paragraph{Per-frame Saturation Ratio Definition}
For a rendered RGB frame at time $t$, $\mathbf{I}_t\in[0,1]^{H\times W\times 3}$, we compute luminance
\begin{equation}
Y_t(\mathbf{p}) \;=\; 0.2126\,R_t(\mathbf{p}) + 0.7152\,G_t(\mathbf{p}) + 0.0722\,B_t(\mathbf{p}),
\end{equation}
where $\mathbf{p}$ indexes pixels.
Given a saturation threshold $\tau=0.98$, the per-frame saturation ratio is
\begin{equation}
\mathrm{SatRatio}_t
\;=\;
\frac{1}{HW}\sum_{\mathbf{p}}\mathbb{I}\!\left[\,Y_t(\mathbf{p}) \ge \tau\,\right].
\end{equation}
We summarize the series $\{\mathrm{SatRatio}_t\}_{t=1}^{T}$ by its mean, standard deviation, and range (max--min):
\begin{equation}
\Big(\mu(\mathrm{SatRatio}),\,\sigma(\mathrm{SatRatio}),\,\mathrm{rng}(\mathrm{SatRatio})\Big).
\end{equation}

\paragraph{Implementation Notes.}
We compute SatRatio directly on the rendered frames produced by each pipeline at $1\times\Delta t$.
All frames are assumed to be normalized to $[0,1]$ per channel before luminance computation.
We use a fixed threshold $\tau=0.98$ to capture near-saturated (blown-out) pixels in a consistent manner across methods and scenes.

\paragraph{Additional Discussion.}
The mean Sat.\% reflects the overall extent of over-exposed regions, while std/range describe the temporal variability of the \emph{near-saturated fraction}.
When the mean level is already low, a relatively larger std/range typically corresponds to local threshold crossings around $\tau$ (e.g., thin structures, specular highlights, or high-contrast boundaries during motion) rather than large-area exposure failures.
Conversely, a ``high-mean, near-zero-variance'' pattern indicates persistent saturation and is characteristic of saturation-dominated degenerate outputs.

\newpage
% =========================
% Ablations (Appendix)
% =========================
\newpage
\section{Ablations}
\label{App:Ablations}

In this subsection, we conduct an ablation study on the implicit MPM solver of \textit{i-PhysGaussian}, focusing on how key numerical strategies within the Newton--GMRES framework affect the trade-off between stability and efficiency. Specifically, we ablate (1) the backtracking line search used in the Newton updates, and (2) the Eisenstat--Walker (EW) forcing term that adaptively controls the accuracy of the inner GMRES solves (i.e., tolerance scheduling).

\paragraph{Evaluation protocol and implementation note.}
We report three solver-trace-derived diagnostics: \textbf{stability} (\textbf{Success}), \textbf{efficiency} (\textbf{Speedup}), and \textbf{convergence quality} (\textbf{RelEnd}), as listed in the corresponding tables.
A \emph{frame} is deemed successful only if \emph{all} substeps within that frame converge.
In our implementation, the number of substeps per frame is an integer
\begin{equation}
\texttt{step\_per\_frame}=\mathrm{round}\!\left(\frac{\texttt{frame\_dt}}{\texttt{substep\_dt}}\right),
\end{equation}
so the effective simulated time interval per frame is $\texttt{step\_per\_frame}\cdot\texttt{substep\_dt}$ (which may differ slightly from \texttt{frame\_dt} when the ratio is not an integer). We apply the same convention consistently across all ablated variants, and all frame-level metrics are computed at the exported frame indices.

% ============================================================
\subsection{Ablation 1: W/O Line Search}
\label{App:AblNoLS}

The backtracking line search serves as a step-size controller for the Newton update. Given the search direction returned by the inner GMRES solve, we first try a full step ($\alpha=1$) and check whether it decreases the Newton residual (or the corresponding objective). If the residual fails to decrease, we backtrack and shrink $\alpha\in(0,1)$ until a sufficient decrease condition is met. Disabling line search removes these extra evaluations and can speed up the loop in some cases, but at the cost of reduced robustness, especially on strongly non-linear or stiff scenes where full Newton steps are more likely to overshoot.

\begin{table}[t]
\caption{Line-search ablation on the \textit{i-PhysGaussian} implicit MPM solver.
\textbf{Success} is the \emph{frame-level} success rate (a frame is successful only if all substeps converge).
\textbf{Speedup} is computed as $\sum t_{\text{Newton}}^{\text{base}}/\sum t_{\text{Newton}}^{\text{noLS}}$.
\textbf{RelEnd} is the mean Newton residual ratio $R_{\text{end}}/R_0$ averaged over \emph{converged substeps}.}
\label{tab:linesearch_ablation}
\centering
\small
\begin{sc}
\setlength{\tabcolsep}{0.5pt}
\begin{tabular}{ccccc}
\toprule
Scene & $\Delta t$ &
\makecell[c]{Success$\uparrow$ \\ \footnotesize base/noLS \\ \footnotesize $\Delta$pp} &
Speedup$\uparrow$ &
\makecell[c]{RelEnd$\downarrow$ \\ \footnotesize base \\ \footnotesize noLS} \\
\midrule

\multirow{3}{*}{\adjustbox{valign=c}{\rotatebox{90}{pillow2sofa\!\!}}} & 1x &
\makecell[c]{63.3 \,/\, 62.6 \\ \textcolor{gray}{\footnotesize $-0.7$}} &
1.02$\times$ &
\makecell[c]{6.28e-4 \\ \footnotesize 6.28e-4} \\

& 10x &
\makecell[c]{71.2 \,/\, 72.4 \\ \textcolor{gray}{\footnotesize $+1.2$}} &
1.00$\times$ &
\makecell[c]{3.09e-4 \\ \footnotesize 2.94e-4} \\

& 20x &
\makecell[c]{66.4 \,/\, 62.6 \\ \textcolor{gray}{\footnotesize $-3.8$}} &
0.99$\times$ &
\makecell[c]{6.20e-4 \\ \footnotesize 5.53e-4} \\

\midrule

\multirow{3}{*}{\adjustbox{valign=c}{\rotatebox{90}{bread\ \ \ \ }}} & 1x &
\makecell[c]{0.3 \,/\, 0.0 \\ \textcolor{gray}{\footnotesize $-0.3$}} &
1.74$\times$ &
\makecell[c]{1.57e-5 \\ \footnotesize 1.59e-5} \\

& 10x &
\makecell[c]{99.3 \,/\, 0.0 \\ \textcolor{gray}{\footnotesize $-99.3$}} &
0.99$\times$ &
\makecell[c]{4.19e-5 \\ \footnotesize 5.72e-5} \\

& 20x &
\makecell[c]{99.3 \,/\, 0.0 \\ \textcolor{gray}{\footnotesize $-99.3$}} &
1.50$\times$ &
\makecell[c]{5.24e-5 \\ \footnotesize --} \\
\bottomrule
\end{tabular}
\end{sc}
\end{table}

Tab.~\ref{tab:linesearch_ablation} reports \textbf{stability} (Success), \textbf{efficiency} (Speedup), and \textbf{convergence quality} (RelEnd) under the line-search ablation. The results support the intended role of line search: when Newton--GMRES directions are already well-behaved, line search is rarely needed; when the problem becomes strongly non-linear, it can be critical for robustness.

On \textit{pillow2sofa}, removing line search has only a \textbf{minor impact} across time steps: Success changes by at most a few percentage points ($-0.7$pp at $1\times$, $+1.2$pp at $10\times$, and $-3.8$pp at $20\times$), while Speedup stays near $1.0\times$ (0.99--1.02$\times$). Consistently, RelEnd remains in the same range, indicating that the Newton updates are typically ``safe'' on this scene and the line search mainly acts as a safeguard rather than a frequently-triggered stabilizer.

In contrast, the \textit{bread} scene amplifies the importance of line search. While the base solver achieves \textbf{near-perfect success} at $10\times$ and $20\times$ (both 99.3\%), \textbf{noLS collapses to 0.0\%} success under the same settings, indicating that full Newton steps frequently overshoot and lead to divergence without step-size control. The ``--'' RelEnd entry at $20\times$ under noLS further corroborates this: there are essentially no converged substeps from which $R_{\text{end}}/R_0$ can be aggregated. (We note that Speedup becomes less informative when Success collapses, since failing runs may terminate early and distort aggregated runtimes.)

% ============================================================
\subsection{Ablation 2: Fixed forcing term (w/o Eisenstat--Walker)}
\label{App:AblFixedForcing}

Fixed-forcing-term Newton--GMRES uses a \emph{manually chosen} tolerance to decide when the inner GMRES loop terminates. Intuitively, when the Newton residual is large, the GMRES solve does not need to be very accurate---an approximate solution can still provide a useful descent direction. As the iteration approaches convergence, however, GMRES typically must solve more accurately to prevent outer progress from stalling. This stage-dependent accuracy demand motivates the \emph{Eisenstat--Walker (EW)} strategy, which adapts the GMRES stopping criterion based on observed outer residual reduction, automatically balancing efficiency and robustness.

In contrast, enforcing a \emph{fixed} forcing term simplifies the solver but generally requires \emph{scene- and step-dependent tuning}: a tolerance that is ``good'' for one setting may become too loose (under-solving) or too strict (over-solving) elsewhere.

\begin{table}[t]
\caption{Fixed forcing-term ablation for the \textit{i-PhysGaussian} implicit MPM solver (\textbf{base}: Eisenstat--Walker adaptive forcing term; \textbf{fixed}: constant GMRES tolerance shared across scenes).
\textbf{Success} is the \emph{frame-level} success rate (a frame is successful only if all substeps converge).
\textbf{Speedup} is computed as $\sum t_{\text{Newton}}^{\text{base}}/\sum t_{\text{Newton}}^{\text{fixed}}$ (larger is faster for \textbf{fixed}).
\textbf{GMRES iters} reports mean/max GMRES iterations aggregated over \emph{converged} Newton solves.
\textbf{RelEnd} is the mean Newton residual ratio $R_{\text{end}}/R_0$ averaged over \emph{converged substeps}.}
\label{tab:fixed_forcing_ablation}
\centering
\small
\begin{sc}
\setlength{\tabcolsep}{0.5pt}
\begin{tabular}{cccccc}
\toprule
\makecell[c]{\scriptsize Scene} & $\Delta t$ &
\makecell[c]{\scriptsize Success$\uparrow$ \\ \scriptsize base/fixed} &
\makecell[c]{\scriptsize Speedup$\uparrow$} &
\makecell[c]{\scriptsize GMRES iters \\ \scriptsize (mean/max) \\ \scriptsize base/fixed} &
\makecell[c]{\scriptsize RelEnd$\downarrow$ \\ \scriptsize base/fixed} \\
\midrule

\multirow{3}{*}{\adjustbox{valign=c}{\rotatebox{90}{pillow2sofa\!\!}}} & 1x &
\makecell[c]{63.3 \,/\, 63.3} &
1.22$\times$ &
\makecell[c]{7.0 \,/\, 18 \\ \footnotesize 4.0 \,/\, 22} &
\makecell[c]{1.53e-6 \\ \footnotesize 1.55e-6} \\

& 10x &
\makecell[c]{70.3 \,/\, 70.3} &
0.99$\times$ &
\makecell[c]{4.1 \,/\, 18 \\ \footnotesize 4.1 \,/\, 18} &
\makecell[c]{2.66e-5 \\ \footnotesize 1.92e-5} \\

& 20x &
\makecell[c]{71.0 \,/\, 71.0} &
1.01$\times$ &
\makecell[c]{6.1 \,/\, 19 \\ \footnotesize 5.3 \,/\, 32} &
\makecell[c]{8.49e-5 \\ \footnotesize 6.58e-5} \\

\midrule

\multirow{3}{*}{\adjustbox{valign=c}{\rotatebox{90}{bread\ \ \ \ }}} & 1x &
\makecell[c]{0.3 \,/\, 0.3} &
0.99$\times$ &
\makecell[c]{11.0 \,/\, 21 \\ \footnotesize 14.3 \,/\, 46} &
\makecell[c]{6.63e-6 \\ \footnotesize 6.65e-6} \\

& 10x &
\makecell[c]{99.3 \,/\, 99.3} &
1.00$\times$ &
\makecell[c]{14.5 \,/\, 39 \\ \footnotesize 20.0 \,/\, 54} &
\makecell[c]{4.21e-5 \\ \footnotesize 4.22e-5} \\

& 20x &
\makecell[c]{99.3 \,/\, 99.3} &
0.89$\times$ &
\makecell[c]{28.3 \,/\, 57 \\ \footnotesize 44.3 \,/\, 84} &
\makecell[c]{5.26e-5 \\ \footnotesize 5.24e-5} \\
\bottomrule
\end{tabular}
\end{sc}
\end{table}

Tab.~\ref{tab:fixed_forcing_ablation} compares \emph{base} (EW adaptive forcing) against \emph{fixed} forcing in terms of \textbf{stability} (Success), \textbf{efficiency} (Speedup), and \textbf{inner-solve workload/convergence quality} (GMRES iters and RelEnd). Overall, the two variants do not exhibit a clear ``converge vs.\ diverge'' split on the reported scenes, since the \textbf{frame-level success rates remain identical} across all settings. However, the fixed strategy shows noticeably higher \emph{sensitivity} in computational behavior: depending on the scene and time step, it may either reduce typical GMRES work or introduce heavier tails (larger worst-case iterations), which is precisely the regime EW aims to handle by adapting inner accuracy to outer progress.

On \textit{pillow2sofa}, Success is \textbf{unchanged} between base and fixed at all time steps (63.3\% at $1\times$, 70.3\% at $10\times$, and 71.0\% at $20\times$). In terms of efficiency, fixed forcing achieves a clear gain at $1\times\Delta t$ with \textbf{Speedup $=1.22\times$}. This aligns with reduced \emph{average} GMRES effort: the mean GMRES iterations drop from \textbf{7.0} (base) to \textbf{4.0} (fixed), while RelEnd remains essentially the same (1.53e-6 vs.\ 1.55e-6). The benefit is not uniform across time steps: at $10\times$ and $20\times$, Speedup becomes marginal (0.99$\times$ and 1.01$\times$), and the iteration profile becomes more mixed. Notably at $20\times$, while the \emph{mean} GMRES iters slightly decreases (6.1 $\rightarrow$ 5.3), the \emph{max} increases substantially (19 $\rightarrow$ 32), indicating that a single fixed tolerance can reduce typical workload but may also yield heavier-tail inner solves in more difficult phases. Across all pillow2sofa settings, RelEnd stays within the same order of magnitude, suggesting that these shifts mainly reflect tolerance sensitivity rather than a systematic change in outer convergence quality.

On the more challenging \textit{bread} scene, the sensitivity of fixed forcing becomes more pronounced in compute cost. Again, Success remains identical between base and fixed (0.3\% at $1\times$, 99.3\% at $10\times$ and $20\times$), and RelEnd is nearly unchanged, indicating comparable convergence quality on converged substeps. However, the GMRES workload increases consistently under fixed forcing, especially at larger time steps: at $1\times$, the mean/max GMRES iters increase from \textbf{11.0/21} to \textbf{14.3/46}; at $10\times$, from \textbf{14.5/39} to \textbf{20.0/54}; and at $20\times$, from \textbf{28.3/57} to \textbf{44.3/84}. Correspondingly, efficiency degrades at $20\times$ with \textbf{Speedup $=0.89\times$} (i.e., a slowdown). Since Success and RelEnd remain essentially unchanged, this pattern is most consistent with \textbf{over-solving}: fixed forcing spends substantially more GMRES iterations to meet a constant inner residual target, yet this extra inner accuracy does not translate into a meaningfully better outer convergence outcome. Taken together, these results motivate our choice of Eisenstat--Walker: it reduces per-scene manual tuning and helps avoid both under-solving and over-solving when rolling out difficult scenes under large time steps.

%%%%%%%%%%%%%%%%%%%%%%%%%%%%%%%%%%%%%%%%%%%%%%%%%%%%%%%%%%%%%%%%%%%%%%%%%%%%%%%
%%%%%%%%%%%%%%%%%%%%%%%%%%%%%%%%%%%%%%%%%%%%%%%%%%%%%%%%%%%%%%%%%%%%%%%%%%%%%%%

\end{document}